%% file: main.tex
\documentclass[10pt,twocolumn,letterpaper]{article}

\usepackage[pagenumbers]{cvpr}

\input{preamble}

\definecolor{cvprblue}{rgb}{0.21,0.49,0.74}
\usepackage[pagebackref,breaklinks,colorlinks,allcolors=cvprblue]{hyperref}

\def\paperID{15609} 
\def\confName{CVPR}
\def\confYear{2025}

\title{Efficient Data Driven Mixture-of-Expert Extraction from Trained Networks}

\author{Uranik Berisha\printfnsymbol{1}\printfnsymbol{2}, Jens Mehnert\printfnsymbol{1} and Alexandru Paul Condurache\printfnsymbol{1}\printfnsymbol{2}\\
	{\small \printfnsymbol{1}Automated Driving Research, Robert Bosch GmbH, 70469 Stuttgart, Germany }\\
	{\small \printfnsymbol{2}Institute for Signal Processing, University of L{\"u}beck, 23562 L{\"u}beck, Germany}\\
	{\tt\small \{Uranik.Berisha,JensEricMarkus.Mehnert,AlexandruPaul.Condurache\}@de.bosch.com}
}

\begin{document}
\maketitle
\input{0_abstract}    
\input{1_introduction}
\input{2_main}

\input{3_conclusion}
{
    \small
    \bibliographystyle{ieeenat_fullname}
    \bibliography{main}
}

 \input{X_suppl}

\end{document}

%% file: preamble.tex
\newcommand{\cmark}{\textcolor{green}{\ding{51}}}  
\newcommand{\xmark}{\textcolor{red}{\ding{55}}}   

\usepackage{colortbl}

\usepackage[acronym]{glossaries}
\loadglsentries{acronyms}

\glsaddall
\glsdisablehyper

\usepackage{amsmath}
\usepackage{amssymb}

\usepackage{pifont}

\usepackage{tabularx} 
\usepackage{booktabs}

\usepackage{graphicx}
\usepackage{float}

\usepackage{subcaption}

\usepackage{siunitx}

\usepackage{textcomp}

\usepackage[symbol]{footmisc}
\newcommand{\printfnsymbol}[1]{\textsuperscript{#1}}

\usepackage{csquotes}

\usepackage{arydshln}

%% file: acronyms.tex
\newacronym[longplural={Mixture-of-Experts}]{moe}{MoE}{Mixture-of-Experts}
\newacronym{dl}{DL}{Deep Learning}
\newacronym{nlp}{NLP}{Natural Language Processing}
\newacronym{cv}{CV}{Computer Vision}
\newacronym{vit}{ViT}{Vision Transformer}
\newacronym{cnn}{CNN}{Convolutional Neural Network}
\newacronym{mhsa}{MHSA}{Multi-Head Self-Attention}
\newacronym{mlp}{MLP}{Multi-Layer Perceptron}
\newacronym[longplural={Mixture-of-Extracted-Experts}]{moee}{MoEE}{Mixture-of-Extracted-Experts}
\newacronym{hdbscan}{HDBSCAN}{Hierarchical Density Based Spatial Clustering for Applications with Noise}
\newacronym{pca}{PCA}{Principal Component Analysis}
\newglossaryentry{encblock}{name={encoder block}, description={encoder block}}
\newglossaryentry{imagenet}{name={ImageNet-1k}, description={ImageNet-1k}}
\newglossaryentry{linlayer}{name={linear layer}, description={linear layer}}
\newglossaryentry{layernorm}{name={layer normalization}, description={layer normalization}}
\newglossaryentry{actfunc}{name={activation function}, description={activation function}}

%% file: 0_abstract.tex
\begin{abstract}
	
	\glspl{vit} have emerged as the state-of-the-art models in various \gls{cv} tasks, but their high computational and resource demands pose significant challenges. While \gls{moe} can make these models more efficient, they often require costly retraining or even training from scratch. Recent developments aim to reduce these computational costs by leveraging pretrained networks. These have been shown to produce sparse activation patterns in the \glspl{mlp} of the encoder blocks, allowing for conditional activation of only relevant subnetworks for each sample.
	
	Building on this idea, we propose a new method to construct \gls{moe} variants from pretrained models. Our approach extracts expert subnetworks from the model's \gls{mlp} layers post-training in two phases. First, we cluster output activations to identify distinct activation patterns. In the second phase, we use these clusters to extract the corresponding subnetworks responsible for producing them. On \gls{imagenet} recognition tasks, we demonstrate that these extracted experts can perform surprisingly well out of the box and require only minimal fine-tuning to regain $98\%$ of the original performance, all while reducing MACs and model size, by up to $36\%$ and $32\%$ respectively.
	
\end{abstract}

%% file: 1_introduction.tex
\begin{figure*}[t]
	\centering
	\begin{subfigure}[t]{0.32\textwidth} 
		\centering
		\includegraphics[width=\textwidth]{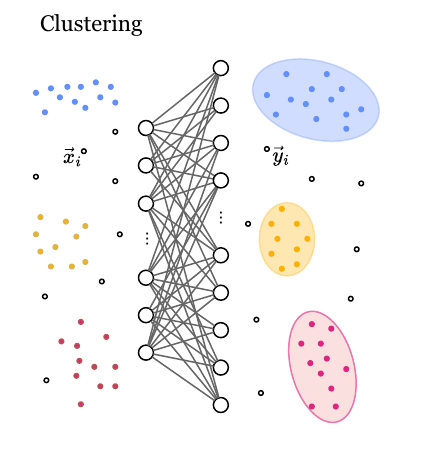}
		\caption{Activations in the hidden layer of the \gls{mlp} $\vec{y}_i$ are clustered using HDBSCAN, which automatically selects the number of clusters.}
		\label{fig:clustering}
	\end{subfigure}
	\hfill
	\begin{subfigure}[t]{0.32\textwidth} 
		\centering
		\includegraphics[width=\textwidth]{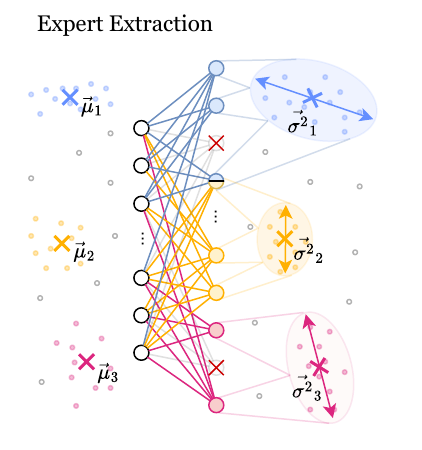}
		\caption{1. Hidden neurons with the highest variance within clusters $\vec{\sigma^2}_c$ are extracted. 2. Clusters are mapped back to the input space and averaged to produce a mean input token $\vec{\mu}_c$ for each expert.}
		\label{fig:extraction}
	\end{subfigure}
	\hfill
	\begin{subfigure}[t]{0.32\textwidth} 
		\centering
		\includegraphics[width=\textwidth]{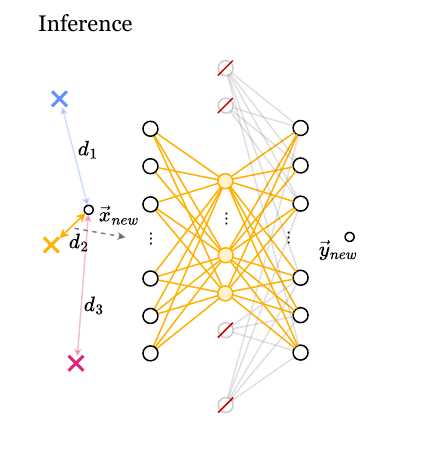}
		\caption{Incoming tokens $\vec{x}_{new}$ are routed based on similarity $d_c$ to the mean input tokens $\vec{\mu}_c$. The selected expert is constructed by removing the hidden neurons not extracted in the expert extraction.}
		\label{fig:inference}
	\end{subfigure}
	\caption{Illustration of the expert extraction process. The process begins with the clustering of activations (a), followed by the extraction of experts from these clusters (b), and finishes with the inference stage, using the extracted experts (c).}
	\label{fig:expert-extraction}
\end{figure*}

\section{INTRODUCTION}

\glspl{cnn} have long dominated \gls{cv} tasks such as image recognition \citep{aiiw1wtfiras}. However, recent advancements have increasingly shifted towards transformers as the base architecture \citep{aiiw1wtfiras, tdeit&dta}. These models have achieved remarkable results but at a significant computational cost \citep{aiiw1wtfiras}. 
Approaches, such as \gls{moe} methods \citep{svwsmoe, mvmsdvtvsmoe, amamtvmoe}, address these computational demands during inference but require expensive and complex training procedures, including additional load-balancing loss terms to ensure balanced expert selection \citep{mmoevtfemtlwmacd}.
Moreover, these models must be trained from scratch on massive datasets like ImageNet-21k, containing over 14 million samples, or Google's JFT-300M dataset, comprising 300 million samples \citep{aiiw1wtfiras}. 
Other solutions, such as DeiT \citep{tdeit&dta}, aim to mitigate the dependency on large datasets by employing strong data augmentations along with regularizations. 

In this work, we approach the problem from a different angle by making use of the vast collection of pretrained networks available. Recent studies have shown that \glspl{mlp} in pretrained language transformers tend to group neurons into subnetworks responsible for distinct functions \citep{emiptt}. For most input, \citet{mtfflamoe} show that the \glspl{mlp} exhibit a high sparsity in their activation patterns, which they use to form \gls{moe} variants of prevalent language transformers.
Since vision transformers must process spatially structured image data, the transferability of methods from language transformers, with their sequential structure, is not trivial. Text tokens can exhibit different activation and modularity patterns due to linguistic rather than spatial structures \cite{aiiw1wtfiras}. 

Building upon these works, our goal is to investigate whether similar phenomenons emerge in \glspl{vit} and leverage them to reduce the size of pretrained \glspl{vit} post-training.
To this end, we introduce a novel expert extraction method that determines the number and size of the experts from the data. This grounds expert configurations in data-driven insights rather than manual selection, enhancing the robustness of the method.
Our expert extraction method consists of two phases. First, we cluster the hidden activations in the \glspl{mlp} using the \gls{hdbscan} algorithm \citep{dbcbohde}. We choose this algorithm, as it allows us to find the number and shape of the experts from the data. In the second phase, we use these clusters to extract subnetworks from the \glspl{linlayer}, prioritizing components based on their variance within the activation patterns. During inference, a simple and efficient routing mechanism, derived from the similarity to the input cluster mean of the corresponding subnetwork, allows us to conditionally select the appropriate expert for processing. 

This extraction process, performed once post-training, generates a \gls{moe} variant of the model, which is why we refer to our method as \enquote{expert extraction}.
\\
\textbf{Our contributions are summarized as follows:}
\begin{itemize}
	\item We introduce a simple yet effective and efficient method for extracting experts from pretrained models, allowing their conversion into \gls{moe} variants.
	\item We apply our method to \glspl{vit} and demonstrate that we can reduce the computational cost and memory footprint while preserving model performance.
	\item We explore the behavior of our method and the resulting \gls{moe} model on different inputs, to find general structural tendencies of emerging substructures and the specialization in their \gls{mlp} layers.
\end{itemize}

\section{RELATED WORK}

\subsection{Transformers}

The transformer architecture, first introduced by \citet{aiayn}, revolutionized \gls{nlp} by employing self-attention mechanisms to capture long-range dependencies. This architecture was later adapted for \gls{cv} by \citet{aiiw1wtfiras} with the introduction of \glspl{vit}, where images are split into patches that serve as tokens. Although \glspl{vit}  achieve state-of-the-art performance, their success depends on the availability of very large datasets and substantial computational resources, making them challenging to train \citep{tdeit&dta}. To address this issue of high data dependency, \citet{tdeit&dta} proposed a series of data augmentation, regularization, and distillation techniques to reduce reliance on large datasets. Despite these innovations, the computational demands of \glspl{vit} remain a bottleneck for practical deployment, particularly on resource-constrained hardware \citep{vtp, w&dpfvt}. \\
An emerging research direction involves understanding how transformers store knowledge \citep{tfflakvm}, whether this knowledge can be modified \citep{efkilm}, and consequently, can be extracted to conditionally activate only subnetworks, to reduce computational costs during inference. \citep{mtfflamoe}

\subsection{Model Compression}

To make \glspl{vit} more practical for deployment on limited hardware, model compression techniques are essential. Pruning methods, which reduce model size by eliminating less important components, remain among the most prevalent approaches \citep{vtp, w&dpfvt, devtwdts, evsftefdvt}. Another promising direction for model compression is clustering-based methods, where clustering algorithms are used to group tokens that can be processed together, improving computational efficiency without sacrificing performance \citep{pvlptcaivt, elsvtfdpwft, ccaauvl}.

While we also make use of clustering algorithms, our work is most closely related to dynamic inference \citep{moddacitblm, dfdcocfte, cmod}. More specifically \glspl{moe}, which divide the model into specialized experts and dynamically route inputs to these experts, thus reducing computational costs while maintaining a large number of parameters \citep{gsgmwccaas}. Inspired by \citet{amole}, \glspl{moe} were successfully extended to \gls{dl} by \citet{lfriadmoe}, scaling \glspl{moe} for deep models. This led to significant advancements in both \gls{nlp} \citep{olnntsgmoel, ststtpmwsaes} and \gls{cv} \citep{svwsmoe, mvmsdvtvsmoe, amamtvmoe}, where \glspl{moe} have proven particularly effective in transformer-based architectures, enabling efficient scaling \citep{svwsmoe}.
However, conventional \glspl{moe} typically define experts before training, which introduces complexities in the training process. Ensuring balanced expert selection requires additional load-balancing loss terms \citep{mmoevtfemtlwmacd}. 
\\
Avoiding this, \citet{mtfflamoe} introduced a novel technique for extracting experts post-training in transformer-based large language models, which aligns with our work.

\subsection{Our Approach}

Our work builds upon the research on emerging subnetworks in pretrained language transformers \citep{mtfflamoe}, which addresses the limitations of traditional \glspl{moe} by enabling the extraction of experts post-training. This offers a more flexible approach to training \glspl{moe} through standard training procedures or even to omit the training step alltogether by using pretrained models.
We investigate these findings and explore these emerging structures within the context of vision. In contrast to \citet{mtfflamoe}, we employ a data-driven approach to the expert configuration, extracting only expert structures that naturally arise in different layers. To do so, we group activations with similar patterns across layers and thereby identify possible subnetworks based on the variance of these activation clusters.

%% file: 2_main.tex
\section{METHODOLOGY}

In this section, we introduce our method for extracting experts and explain how these experts are structured, represented, and used during inference.
As is common in \gls{moe} literature, the \glspl{mlp} are partitioned into experts that operate on the token level. 
We therefore employ a token-wise clustering of the activations in the hidden \glspl{linlayer} to identify subnetworks, which can then dynamically process individual tokens, through efficient routings.
\\
Our method consists of three major steps:
\begin{enumerate}
	\item \textbf{Activation Clustering} The layer activations are clustered over multiple batches (\cref{fig:clustering}).
	\item \textbf{Expert Extraction} For each cluster, a subnetwork is identified based on the cluster variance (\cref{fig:extraction}).
	\item \textbf{Inference} The tokens are routed to the subnetworks based on the clusters to generate an expert-specific output (\cref{fig:inference}).
\end{enumerate}

\subsection{Step 1: Activation Clustering}

First, we need to identify clusters of similar activations. To achieve this, we assume that a randomly sampled subset of the data can sufficiently represent the underlying activation patterns (see \Cref{effect-of-sample-size} for further details). Based on this assumption, we record the activations \( \mathbf{y}_i \) of individual tokens \( \mathbf{x}_i \) in each hidden layer, which are then clustered.

While our method does not require a specific clustering algorithm, it must enable a data-driven estimation that avoids restrictive assumptions about the number of clusters \( k \) or potential activation patterns, as we want to extract the number and shape of the experts from the data.
Given the high dimensionality \( e \) of the embedding space, the clustering algorithm must also be computationally efficient and be able to handle noise.
\\
An algorithm fulfilling these constraints is \gls{hdbscan}, a robust clustering algorithm well-suited for high-dimensional data. It has one primary hyperparameter, \textit{minimum cluster size}, which acts like a blurring filter by determining the minimum number of points needed to form an individual cluster. This merges smaller noise clusters into larger, more significant ones and thus controls the granularity of the experts. It works well across a range of values, which makes it easy to determine (see \Cref{effect-of-hyperparameters} for further details). \gls{hdbscan}'s flexibility allows us to apply it across all layers. This avoids manual tuning of further hyperparameters, as it has only data-dependent decisions on where specialization patterns emerge.
This means that the number of experts in a layer \( l \) is given by the number of clusters detected \( k_l \). If no clusters are detected in a layer, we assume that the \gls{linlayer} has not yet formed specialized subnetworks and is instead processing general features, so we leave the layer unchanged.

\subsection{Step 2: Expert Extraction}

Given these activation clusters, our goal is to form the desired experts by extracting specialized subnetworks. This requires representing these clusters in a way that facilitates both subnetwork extraction and an efficient comparison of new inputs with existing clusters, to allow effective input routing.

\subsubsection{Expert Representation}

We use simple descriptive statistics, specifically the variance \(\vec{\sigma^2}_c\) of the activation vectors for each cluster \( c \), to identify which components are most critical. Activations with higher variances within the clusters capture more diverse patterns, while those with lower variance are assumed to contribute less to the specificity of the cluster (see \Cref{variance-vs-magnitude} for further details). 

After prioritizing activation components based on their variance, we extract the experts by selecting the subset of hidden neurons within the MLP that produced these components, thus identifying parts of the network that are most important for representing the clusters. This corresponds to choosing columns in the weight matrix \( \mathbf{W}^{(1)} \). 
Importantly, experts are extracted independently, allowing them to overlap and not necessarily cover all weights. This independent extraction enables the removal of neurons not selected by any expert, reducing the parameter count and minimizing the memory footprint.
To further optimize memory, weights for each expert are shared across the layer, eliminating the need to store redundant parameters.

Inspired by other variance-based algorithms, such as \gls{pca}, we define a hyperparameter \textit{extraction percentage} \( p\% \), which determines how much of the total variance the cumulative variance of selected output neurons must cover. 
\\
This parameter reflects how much information we want to keep, allowing adjustment of the model's level of detail (see \Cref{effect-of-hyperparameters} for further details).

\subsubsection{Input Cluster Representation}

Because linear transformations preserve clusters, we can map clusters from the activation space back to their respective inputs, thereby forming input clusters. To assess which cluster a new input \( \vec{x}_{new} \) belongs to, several methods can be used, such as explicit density estimations or \( k \)-nearest neighbors. However, these approaches are resource-intensive, which may be impractical in real-time inference or on resource-constrained hardware.
Instead, we make a simplifying assumption that the clusters are spherical. This allows us to represent each cluster by its mean input vector \( \vec{\mu}_c \), similar to the K-Means algorithm. We then compute the mean input vector for each cluster, which we use for efficient routing of new inputs.
We make this simplifying assumption in favor of a faster compute, by relying on the flexibility of the neural network to compensate for minor errors during the fine-tuning phase, where the model can adjust its parameters to correct routing inaccuracies.

\subsection{Step 3: Inference}

With the precomputed mean input vectors \( \vec{\mu}_c \) and the extracted sub-networks for each cluster, we now describe how the model operates during inference. The process involves routing tokens to the appropriate expert and generating expert-specific outputs.

\subsubsection{Expert Routing}

For each token, we calculate the pairwise cosine similarity \( d_c \) (see \Cref{cosine-vs-euclidian} for further details) between the input token vector \( \vec{x}_{new} \) and the mean input vectors of all experts \( \vec{\mu}_c \). The cluster with the highest cosine similarity is then selected as the most appropriate match, and the token is routed to the corresponding expert. This token-wise routing ensures that each token is processed by the expert most specialized for handling its features.

\subsubsection{Expert Output}

Once routed to an expert, the token is processed by a subset of weights from the original \gls{linlayer}, selected using binary masks \( \mathbf{M} \) that correspond to columns of the weight matrix \( \mathbf{W}^{(1)} \). Since each expert uses fewer neurons than the full \gls{linlayer}, the output may not match the original embedding dimension \( e \). To address this mismatch, the input layer of the second \gls{linlayer} is reduced accordingly by removing the corresponding rows of the weight matrix \( \mathbf{W}^{(2)} \), using the transposed binary mask \( \mathbf{M}^{T} \). This ensures that outputs from different experts are compatible and further reduces the computational load.

\subsubsection{Computational Efficiency}

Our approach saves resources by removing neurons that are unused by any expert and by reducing the number of hidden neurons computed for each token. Although the routing mechanism introduces a small overhead due to the matrix multiplication required for calculating the pairwise cosine similarity, this overhead is negligible compared to the savings gained from reducing the size of computed activations. Specifically, for \(k\) experts, we perform dot products with cluster means of dimension \(e\), resulting in a \(k \times e\) matrix multiplication with each incoming token. In contrast, the upward projection from \(e\) to a hidden representation of size \(3e\), amounts to a \(3e \times e\) matrix multiplication per token. Because \(k \approx 10 << 3e \) (see Appendix \Cref{tab:nr-experts}), the savings significantly outweigh the additional routing cost.

\begin{table*}[t]  
	\centering
\begin{tabularx}{\textwidth}{
		>{\arraybackslash}X  
		>{\centering\arraybackslash}X
		>{\centering\arraybackslash}X
		>{\centering\arraybackslash}X
		>{\centering\arraybackslash}X
	}
	\toprule
	\textbf{Model}         & \textbf{MACs (G)} & \textbf{Parameters (M)} & \textbf{Acc. Retention (\%)} & \textbf{Top-1 Acc. (\%)} \\ 
	\midrule
	\rowcolor{gray!20} DeiT-T         & 1.26  & 5.72  & --          & 72.02   \\
	DeiT-T-MoEfication         & \textbf{0.94 (-27.4\%)}  & 5.72  & \textbf{57.41}          & 68.91   \\
	DeiT-T-MoEE (ours)    & 0.95 (-27.0\%)    & \textbf{4.57 (-20.1\%)} & 55.33          & \textbf{69.73}    \\
	\midrule
	\rowcolor{gray!20} DeiT-S         & 4.61  & 22.05   & --          & 79.70 \\
	DeiT-S-MoEfication         & 3.31 (-29.0\%)  & 22.05   & 67.08          & 77.10 \\ 
	DeiT-S-MoEE (ours)    & \textbf{3.19 (-30.6\%)}    & \textbf{16.54 (-25.0\%)} & \textbf{67.60}          & \textbf{78.11}    \\
	\midrule
	\rowcolor{gray!20} DeiT-B         & 17.58 & 86.57   & --          & 81.73 \\
	DeiT-B-MoEfication         & 11.41 (-34.8\%) & 86.57   & 57.20          & 77.63 \\ 
	DeiT-B-MoEE (ours)    & \textbf{11.14 (-36.3\%)}   & \textbf{58.55 (-32.4\%)} & \textbf{68.54}          & \textbf{80.12}    \\
	\bottomrule
\end{tabularx}

\caption{Performance and parameter comparison, evaluating four metrics: \textbf{Accuracy Retention}: retained accuracy after expert extraction, before fine-tuning; \textbf{Top-1 Accuracy}: final accuracy after fine-tuning; \textbf{MACs}: computational operations, measured in billions of operations; and \textbf{Parameters}: the total model size in millions of parameters. Our method (MoEE) achieves competitive accuracy with significant reductions in MACs and parameters, especially in the DeiT-S and DeiT-B models.}

	\label{tab:performance}
\end{table*}

\begin{figure*}[t]
	\centering
	\begin{subfigure}[t]{\columnwidth}
		\centering
		\includegraphics[width=.9\columnwidth]{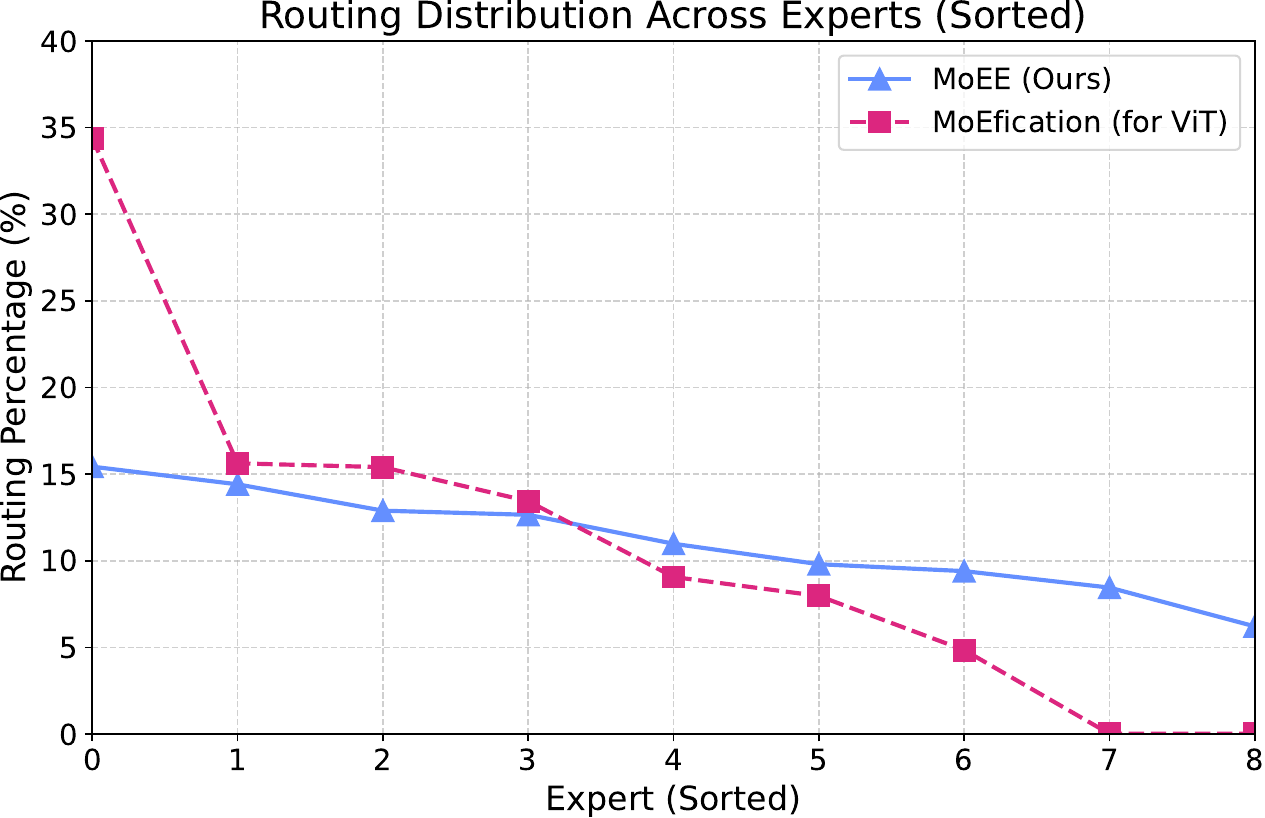}
		\caption{Layer 6}
		\label{fig:routing-balance-6}
	\end{subfigure}
	\hfill
	\begin{subfigure}[t]{\columnwidth}
		\centering
		\includegraphics[width=.9\columnwidth]{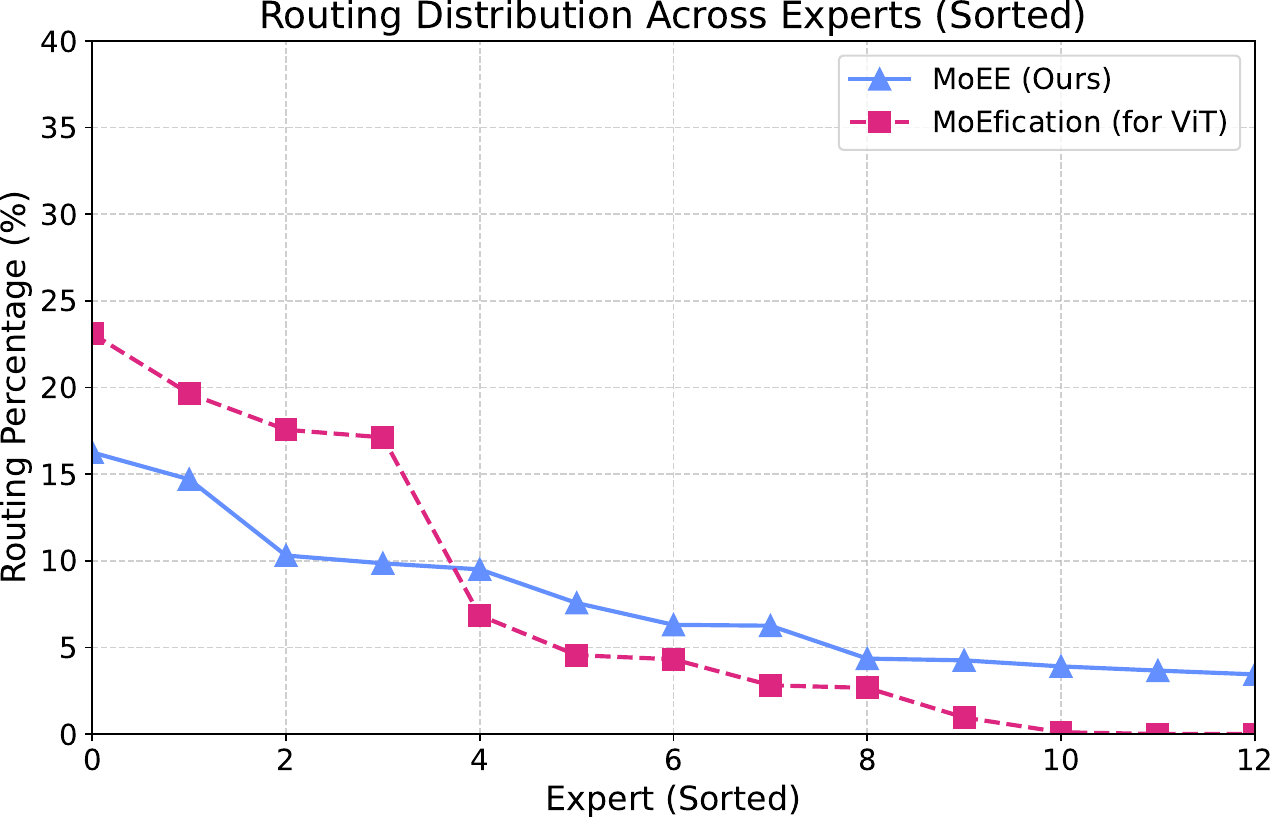}
		\caption{Layer 11}
		\label{fig:routing-balance-11}
	\end{subfigure}
	\caption{Sorted routing distributions across experts for all \gls{imagenet} classes at different layers, demonstrating a more balanced routing load compared to ViT-MoEfication across different layers, effectively reducing the need for additional load-balancing terms.}
	\label{fig:combined-routing-balance}
\end{figure*}

\section{EXPERIMENTS}
\label{sec:experiment}

We apply our expert extraction method to standard vision transformer architectures, namely DeiT-Tiny, DeiT-Small, and DeiT-Base \citep{tdeit&dta}, pretrained on the \gls{imagenet} dataset \citep{ialshid}. This results in their respective \gls{moee} forms: DeiT-T-MoEE, DeiT-S-MoEE, and DeiT-B-MoEE. We compare the results with our reimplementation of \citet{mtfflamoe} for \gls{vit}, denoted by the MoEfication tag in the respective experiments.
Before fine-tuning, we evaluate the performance of these models directly on \gls{imagenet} to measure how much accuracy is retained immediately after the expert extraction. 

The models are then fine-tuned for 30 epochs with a batch size of 32 using knowledge distillation from the corresponding unmodified model. We use the AdamW optimizer \citep{dwdr} with a weight decay of 0.01 and an initial learning rate of 1.5e-5, which is decayed using a cosine annealing scheduler \citep{ssgdwwr}. To demonstrate our commitment to efficient machine learning and to highlight the efficacy of our method, all models are evaluated and fine-tuned on a single NVIDIA Tesla T4 GPU each. 

\subsection{Effect of Hyperparameters}
\label{effect-of-hyperparameters}
Our method introduces two key hyperparameters, namely the \textit{minimum cluster size} and the \textit{extraction percentage}, which can be tuned to extract more or less and bigger or smaller experts respectively. Both of which influence the trade-off between MACs reduction and model accuracy.

When using \gls{hdbscan}, as the \textit{minimum cluster size} increases, the number of distinct experts decreases because variations in token density are increasingly interpreted as noise within larger clusters. This results in less specialized experts, requiring more neurons to represent the cluster. Consequently, increasing the minimum cluster size while keeping the extraction percentage constant leads to a reduction in MACs savings but improves accuracy retention. 
\\
We performed a hyperparameter search, evaluating minimum cluster sizes ranging from $0.05\%$ to $1.5\%$ of the total token sample size for the extraction. We identified $0.6\%$ as a favorable trade-off between MACs reduction and accuracy.

The \textit{extraction percentage} determines how much of the representational capacity of each expert is retained during extraction. Higher percentages result in larger experts in terms of neuron count, leading to better model accuracy but reduced MACs savings. Through a hyperparameter search, we find that an extraction percentage between $70\%$ and $90\%$ offers a good balance between computational savings and accuracy. We use an extraction percentage of $80\%$ for all following experiments (see Appendix \Cref{sec:effect-on-hyperparam} for further details). 

\begin{figure*}[t]
	\centering
	\begin{subfigure}[t]{\columnwidth}
		\centering
		\includegraphics[width=.9\columnwidth]{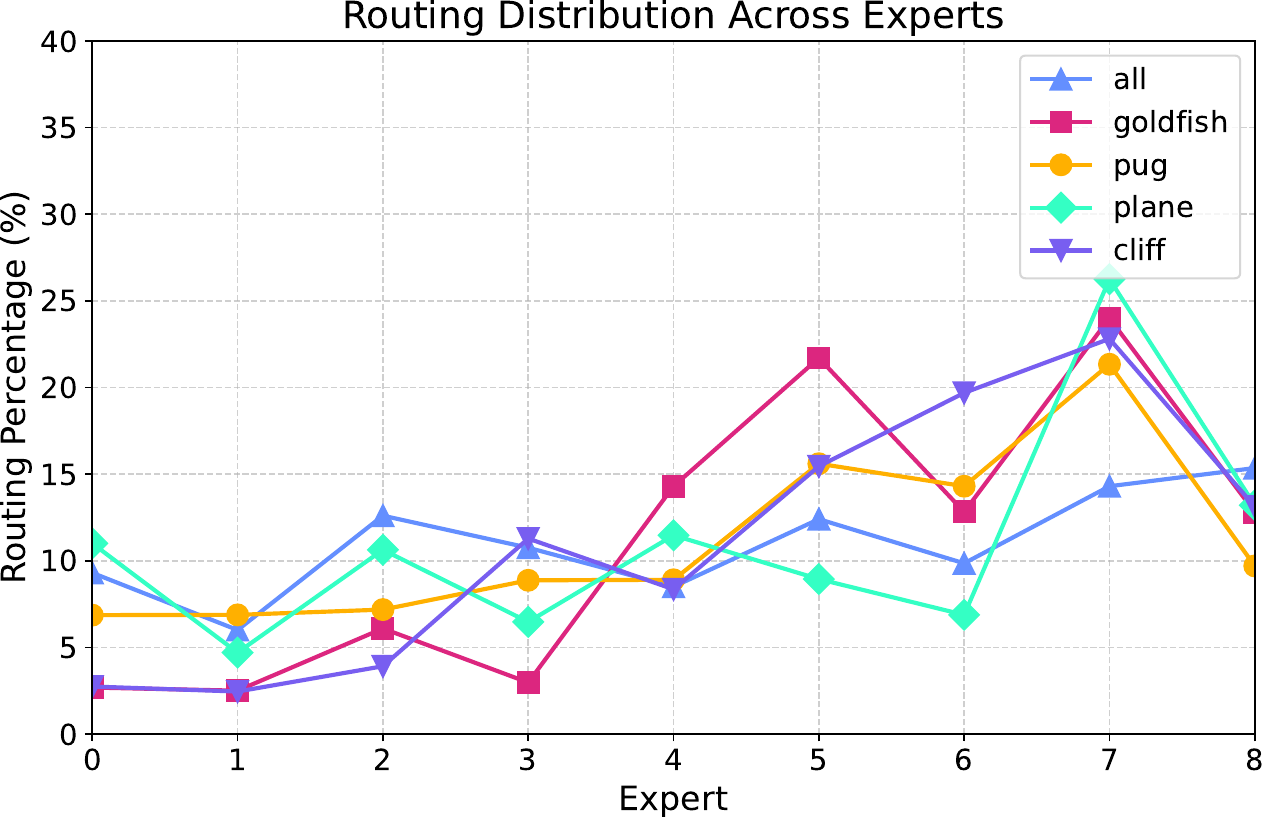}
		\caption{Layer 6}
		\label{fig:random-routings-6}
	\end{subfigure}
	\hfill
	\begin{subfigure}[t]{\columnwidth}
		\centering
		\includegraphics[width=.9\columnwidth]{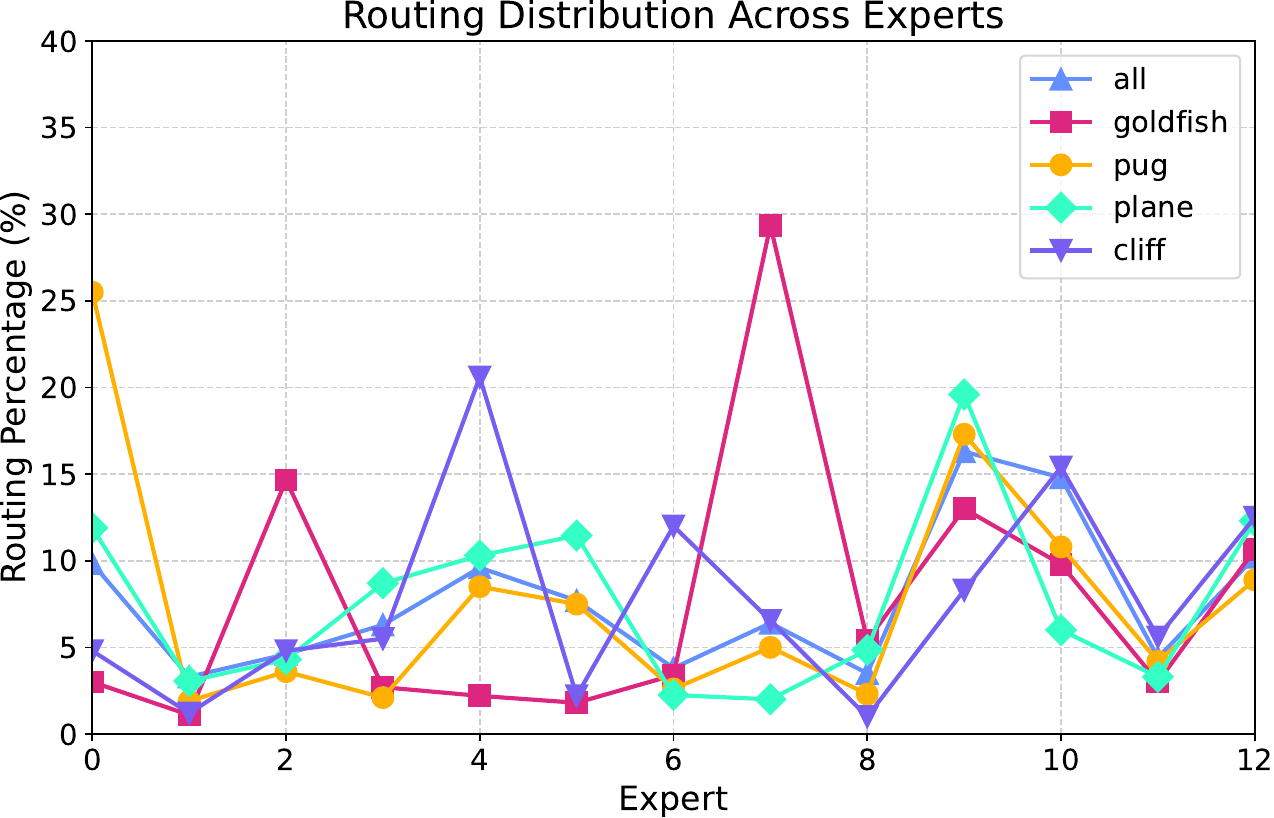}
		\caption{Layer 11}
		\label{fig:random-routings-11}
	\end{subfigure}
	\caption{Token routing distributions at different layers for randomly selected classes (goldfish, pug, plane, and cliff) compared to the distribution across all \gls{imagenet} classes (all). Layer 6 shows a relatively even distribution across experts, indicating less class-specific specialization, Layer 11 shows distinct spikes in the routings, as tokens are more selectively routed based on class-specific features.}
	\label{fig:combined-random-routings}
\end{figure*}

\begin{figure*}[t]
	\centering
	\begin{subfigure}[t]{\columnwidth}
		\centering
		\includegraphics[width=.9\columnwidth]{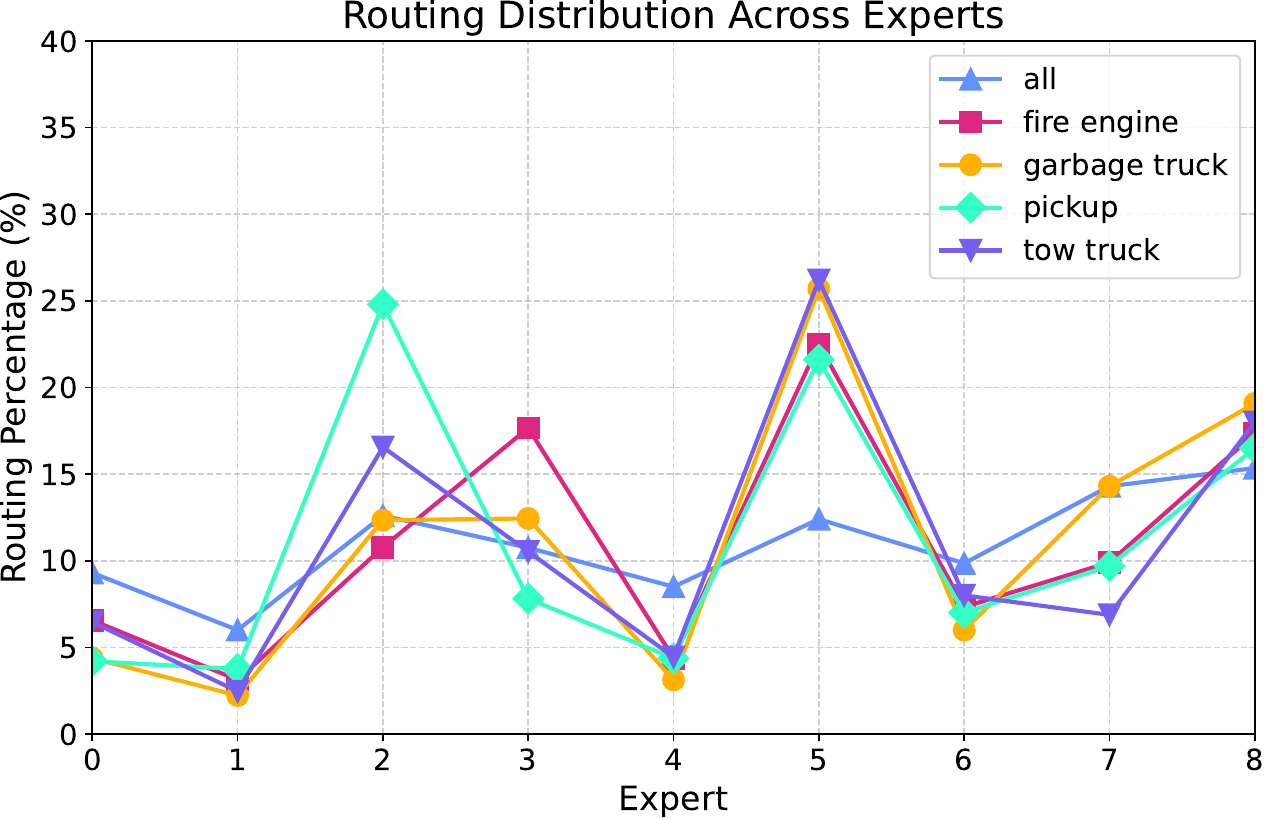}
		\caption{Layer 6}
		\label{fig:truck-routings-6}
	\end{subfigure}
	\hfill
	\begin{subfigure}[t]{\columnwidth}
		\centering
		\includegraphics[width=.9\columnwidth]{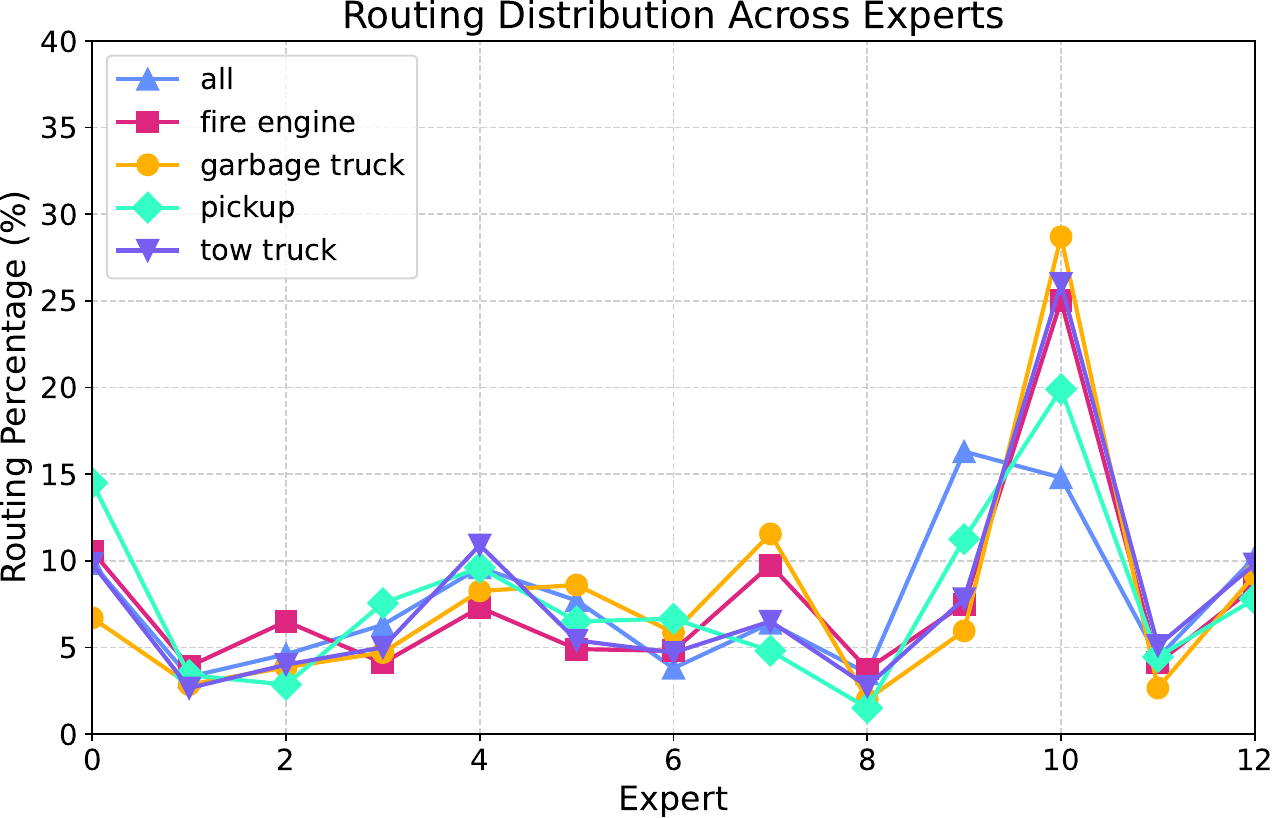}
		\caption{Layer 11}
		\label{fig:truck-routings-11}
	\end{subfigure}
	\caption{Token routing distributions at different layers for visually similar truck-like classes (fire engine, garbage truck, pickup, and tow truck) compared to the distribution across all \gls{imagenet} classes (all). Both layers shows similar distributions across experts for all truck-like classes. This indicates a processing through similar expert selections, reflecting the effectiveness of the routing mechanism.}
	\label{fig:combined-truck-routings}
\end{figure*}

\subsection{Mixture-of-Expert Extraction on ImageNet}

\Cref{tab:performance} shows the main results of our method compared to the baseline DeiT architectures and the MoEfication approach. 
\\
In our evaluation we report the Top-1 Accuracy before finetuning (Acc.\ Retention), the Top-1 Accuracy after finetuning (Top-1 Acc.), the MACs and the parameters. As the two key metrics, we mainly consider the final accuracy of our models after finetuning and the computational savings. 

The expert extraction reduces MACs by $27\%$, $31\%$ and $36\%$, while also reducing the memory footprint by $20\%$, $25\%$ and $32\%$ for the tiny, small and base variants respectively.
Notably, the larger the model, the more effective our method appears to be in both accuracy retention and MACs reduction. For instance, the DeiT-B-MoEE model achieves a significant MACs reduction of $36.3\%$ and a memory reduction of $32.4\%$ while maintaining $84\%$ of the Top-1 Accuracy of the unmodified DeiT-B baseline out of the box, and regaining $98\%$ of the original performance. 
We demonstrate the generalizability of our method to other architectures and dataset by applying the expert extraction on Swin- \citep{sthvtusw} and ConvNeXt Models \citep{acft2} as well as the DeiT models trained on CIFAR-100 \citep{lmloffti} (see Appendix \Cref{sec:generalizability}). 
\\
When compared to MoEfication (see Appendix \Cref{sec:comparison-to-moefication} for an overview of the differences), our approach particularly improves the Top-1 Accuracy. In the DeiT-B variant, DeiT-B-MoEE achieves a $2.5\%$-points higher Top-1 Accuracy with $1.5\%$-points higher MACs reduction, while simulatneously removing $32.4\%$ of the parameters. This suggests that our data-driven expert extraction process is more effective in capturing relevant activation patterns, allowing for an improved trade-off between computational savings and accuracy, particularly for larger vision models.

\Cref{fig:combined-routing-balance} presents the sorted routing distributions across experts for both our method and our implementation of the MoEfication approach adaptet for \glspl{vit}, highlighting differences in load balancing between the two methods. As noted in \citet{mtfflamoe}, the MoEfication approach tends to exhibit an unbalanced routing distribution, with certain experts receiving a disproportionately high percentage of tokens while others remain underutilized. 
\\
This imbalance necessitates additional load-balancing terms in conventional \glspl{moe} to maintain a more even distribution.
In contrast, our method achieves a significantly more balanced distribution, with routing percentages closer to the optimal load of around 11\%, at Layer 6, and around 8\%, in Layer 11. This balanced routing distribution highlights another advantage of our approach, as it avoids the need for supplementary load-balancing mechanisms.

\begin{figure}[t]
	\centering
	\begin{subfigure}[t]{0.48\columnwidth}
		\centering
		\includegraphics[width=\textwidth]{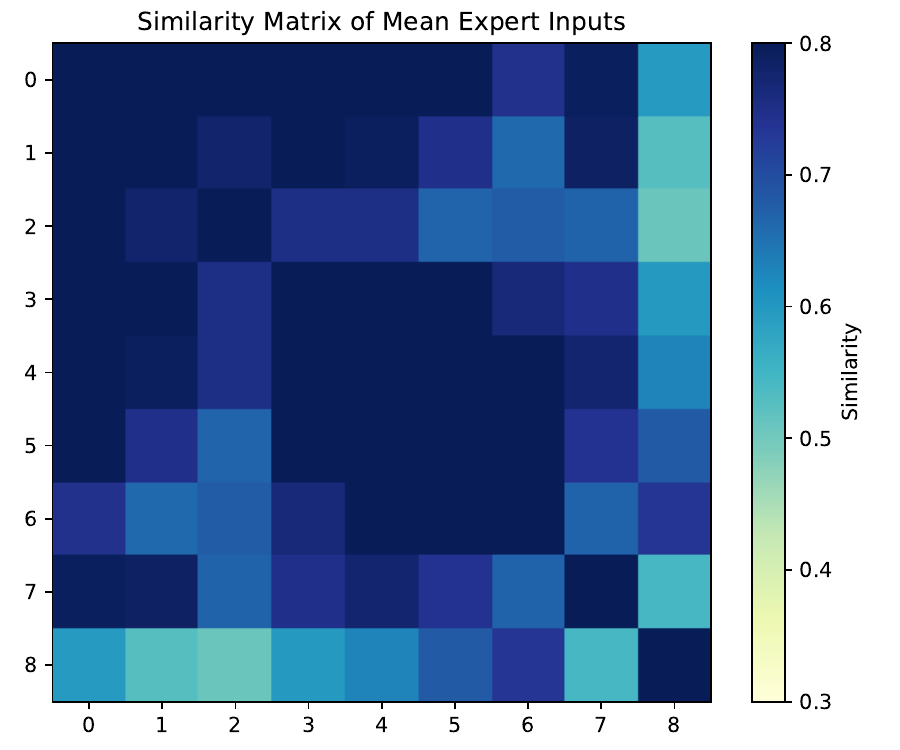}
		\caption{Layer 6}
		\label{fig:similarity-matrix-6}
	\end{subfigure}
	\hfill
	\begin{subfigure}[t]{0.48\columnwidth}
		\centering
		\includegraphics[width=\textwidth]{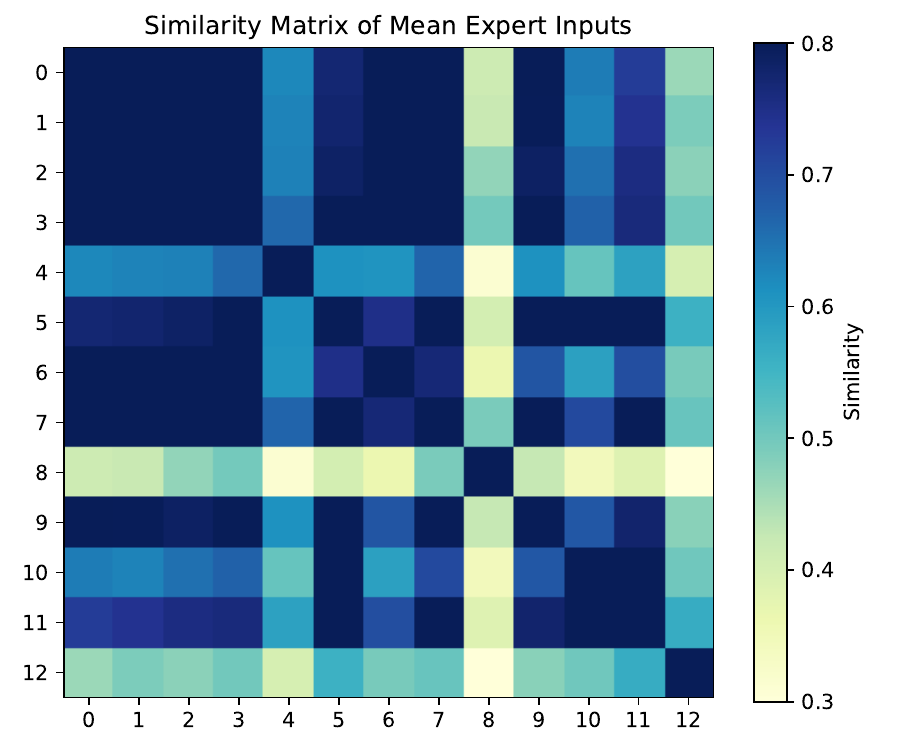}
		\caption{Layer 11}
		\label{fig:similarity-matrix-11}
	\end{subfigure}
	\caption{Similarity matrices of mean expert inputs at different layers, illustrating the relationships between inputs of  different experts. Layer 6 shows high similarities, with several experts' inputs being closely related, indicating less specialization. Layer 11 exhibit distinct differences, with some experts showing near-orthogonal inputs, indicating higher specialization.}
	\label{fig:combined-similarity-matrix}
\end{figure}

\subsection{Expert Routing Analysis}

To better understand the behavior of our approach, we analyze the distribution of tokens routed to each expert at different layer depths. To do this, we count the number of tokens routed to each expert during the evaluation of validation data across individual classes. We then compare randomly selected classes, visually similar classes (selected manually), and the overall routing behavior across all classes to highlight differences in routing patterns.
\\
In this analysis, we use the DeiT-B-MoEE model, where we apply our expert extraction to all 12 layers (indexed 0 to 11). However, since earlier layers of \glspl{vit} capture more general features, as noted by recent studies \cite{dvtslcnn}, \gls{hdbscan} does not detect distinct expert clusters in the initial layers. Instead, we observe distinct expert formations only in layers 6 through 11 (see Appendix \Cref{tab:nr-experts}), as these deeper layers show sufficient variation in activation patterns. 
By examining the first and last extracted expert layers, we identify that experts in deeper layers show more specialized routing behavior, confirming that the network develops increasing specialization with depth, as shown in the following visualizations.

\Cref{fig:combined-random-routings} illustrates the token routing distributions for randomly selected classes: goldfish, pug, plane, and cliff. At Layer 6 (\cref{fig:random-routings-6}), we observe that the routing behaviors do show class specific differences, but still are distributed across the available experts without much variance, suggesting that the earlier layers process these classes more generally.
In contrast, at Layer 11 (\cref{fig:random-routings-11}), the routing patterns become more distinct, with each class exhibiting clear spikes in expert selection. These spikes indicate that the model has learned to differentiate between these classes more effectively at this layer, assigning class-specific processing paths. 

\Cref{fig:combined-truck-routings} shows the token routing distributions for selected truck-like classes: fire engine, garbage truck, pickup, and tow truck. At both layers, the routing patterns across these classes show a high degree of overlap, suggesting that the model processes visually similar classes through similar expert paths in all layers. This consistent routing pattern across related classes reflects the effectiveness of the routing mechanism, even at layers with less specialization.
Notably, the distribution for the truck-like classes closely aligns with the distribution of all 1000 \gls{imagenet} classes.
\\
\\
This contrasts with other groups of visually similar classes, such as shark-like classes (shown in \Cref{appendix:fig:comparison-routings}), which also display a cohesive routing pattern among themselves but differ significantly from the distribution of all classes. This suggests that the truck images contain more generic features that are common across many classes.

\Cref{fig:combined-similarity-matrix} compares the similarity matrices of mean expert inputs. In Layer 6 (\cref{fig:similarity-matrix-6}), the similarity between the mean inputs is relatively high, with several experts (specifically experts 0, 1, 3, 4, and 5) showing closely related mean tokens with a cosine similarity of over $0.8$. This again suggests a lower level of specialization among experts at this earlier layer.
By contrast, the similarity matrix at Layer 11 (\cref{fig:similarity-matrix-11}) reveals a more diverse distribution of experts, with distinct differences between them. Notably, experts 12 and 8 exhibit near-orthogonal relationships with most other experts, indicating a higher degree of specialization.

\section{ABLATION}
\begin{table}[t]  
	\centering
	\begin{tabularx}{\columnwidth}{
			>{\centering\arraybackslash}X  
			>{\centering\arraybackslash}X
			>{\centering\arraybackslash}X
			>{\centering\arraybackslash}X
		}
		\toprule
		\textbf{HDBSCAN Clustering} & \textbf{Variance Extraction} & \textbf{Input Routing} & \textbf{Top-1 Acc. (\%)} \\
		\midrule
		\xmark & \xmark & \xmark & 58.64 \\
		\cmark & \xmark & \xmark & 75.36 \\
		\cmark & \cmark & \xmark & 77.68  \\
		\cmark & \cmark & \cmark & 80.12 \\
		\bottomrule
	\end{tabularx}
	\caption{Ablation study results showing the impact of each introduced component (HDBSCAN clustering, variance-based extraction, and input-based routing) on Top-1 Accuracy. Each row adds a component to demonstrate its effect, with all components together yielding the highest accuracy (80.1\%).}
	\label{tab:ablation-study}
\end{table}

To further evaluate our design choices, we conduct an ablation study and test the assumptions made in the methodology. This provides deeper insights into the effect of these assumptions on model performance.
We systematically remove or replace each component of our methodology with a random counterpart and evaluate the resulting ablated models after the previously described fine-tuning. This includes selecting random clusters, selecting random hidden neurons for the expert extraction, and routing incoming tokens to random experts.

The results of our ablation study, presented in \Cref{tab:ablation-study}, showcase the usefulness of each component in our methodology. When \gls{hdbscan} clustering alone is used without variance-based extraction or input-based routing, the model achieves a Top-1 Accuracy of $75.4\%$, an improvement of $16.8$ percentage points over the baseline with random expert extraction ($58.6\%$). Adding the variance-based neuron selection increases this accuracy by 2.3 percentage points, demonstrating that our variance-based prioritization of neurons contributes significantly to the expert extraction.
\\
Including our input-based routing raises the Top-1 Accuracy by an additional $2.4$ percentage points, demonstrating the effectiveness of routing tokens to the cluster with the most similar input mean. Together, these results confirm the efficacy of each component.

\subsection{Alternative Clustering Methods}
\label{alternative-clustering}
While our method uses HDBSCAN for its ability to extract a variable number of clusters per layer without manual tuning, we also evaluate alternative clustering algorithms. These include DBSCAN and OPTICS as related density-based methods, and K-Means and BIRCH as partition-based baselines. 
We find that density-based methods perform significantly better in accuracy retention as well as final accuracy, with HDBSCAN consistently yielding the best results across all metrics (see Appendix \Cref{sec:effect-of-clustering}).

\subsection{Variance vs. Magnitude-Based Extraction}
\label{variance-vs-magnitude}
In our variance-based approach, components with higher variance are prioritized, assuming they capture more relevant information. An alternative method would be to prioritize based on the magnitude of mean activations, where components with higher mean activations are assumed to be more significant. While both methods select important components, the variance-based approach better captures the diversity in activations, which proved to be more beneficial for expert extraction (see Appendix \Cref{sec:effect-of-extraction-and-routing}).

\subsection{Alternative Routing Methods}
\label{cosine-vs-euclidian}
We also compare the cosine similarity to a pairwise Euclidean distance for expert routing. The expert with the smallest Euclidean distance is selected as the best match. Both routing approaches yield similar end accuracies. However, cosine similarity offers computational efficiency as it can be calculated via a single unnormalized matrix multiplication, since we only need to select the relative distances. This advantage makes cosine similarity our default choice for routing (see Appendix \Cref{sec:effect-of-extraction-and-routing}).

\subsection{Effect of Sample Size on Extraction Stability}
\label{effect-of-sample-size}
To validate our assumption that a randomly sampled subset of the data can sufficiently represent the underlying activation patterns, we vary the number of tokens used for the extraction and look at the number and size of experts extracted by our method. To this end, we find that stability of our method is indeed affected by the number of samples used for extraction. At very low sample sizes, the extraction is highly dependent on the specific samples chosen, leading to variability in expert configurations.
However, starting from approximately 100,000 tokens (which corresponds to around 500 sample images), our method begins to show significant reductions in variance between different sample sets. 
Notably, these 500 images do not even cover all 1,000 \gls{imagenet} classes. We attribute this early stabilization to the presence of common tokens across classes, similar to the routings of truck-like classes, where the tokens follow the routing distribution of all classes (see Appendix \Cref{sec:sample-size}).

%% file: 3_conclusion.tex
\section{DISCUSSION}

Our approach of extracting experts post-training and fine-tuning the model offers two key advantages over traditional methods of constructing \glspl{moe}. Firstly, it enables the creation of \gls{moe} variants from existing \glspl{vit} without the need for retraining from scratch, allowing significant computational savings when working with pretrained models. Secondly, it eliminates the need for specialized training procedures typically associated with \glspl{moe}. By training the base model using standard optimization algorithms and extracting experts afterwards, this method can lead to faster overall training times, as the fine-tuning phase is relatively lightweight compared to full retraining.

Although the accuracy retention after extraction is not sufficient for direct deployment, it is noteworthy that the model retains substantial accuracy given the major structural changes introduced during expert extraction. By conditionally activating specific subnetworks for each input, our approach retains a high capacity for diverse representations, as experts can specialize in processing different input types, while still reducing the computational load per token.
This enables an immediate evaluation of the model and provides a reliable estimate of the accuracy-to-savings trade-off without requiring fine-tuning, allowing for straightforward hyperparameter selection.

\section{CONCLUSION}

We introduce a method for constructing \gls{moe}-variants from pretrained \glspl{vit}, with minimal computational costs compared to training a new \gls{moe} model from scratch. Demonstrated by the experiments, our approach offers significant efficiency gains post-training, while maintaining comparable performance levels. Extensive ablation studies further highlight the effectiveness of our method.

In addition to demonstrating practical gains, our work provides analytical insights into activation clustering, expert formation, and the modularity emerging in ViTs at varying depths. For future work, we plan to explore extensions of this method to other architectures, broadening its applicability and impact, and to investigate potential combinations with pruning and other compression techniques to achieve further gains in efficiency.

We believe the novelty and simplicity of baseing our approach on cluster statistics offers a new perspective on model compression and optimization, and we hope it inspires further developments in the \gls{cv} research community.

%% file: X_suppl.tex
\clearpage
\setcounter{page}{1}
\maketitlesupplementary

\begin{figure}[t]
	\centering
	
	\begin{subfigure}[t]{0.48\textwidth}
		\centering
		\includegraphics[width=\textwidth]{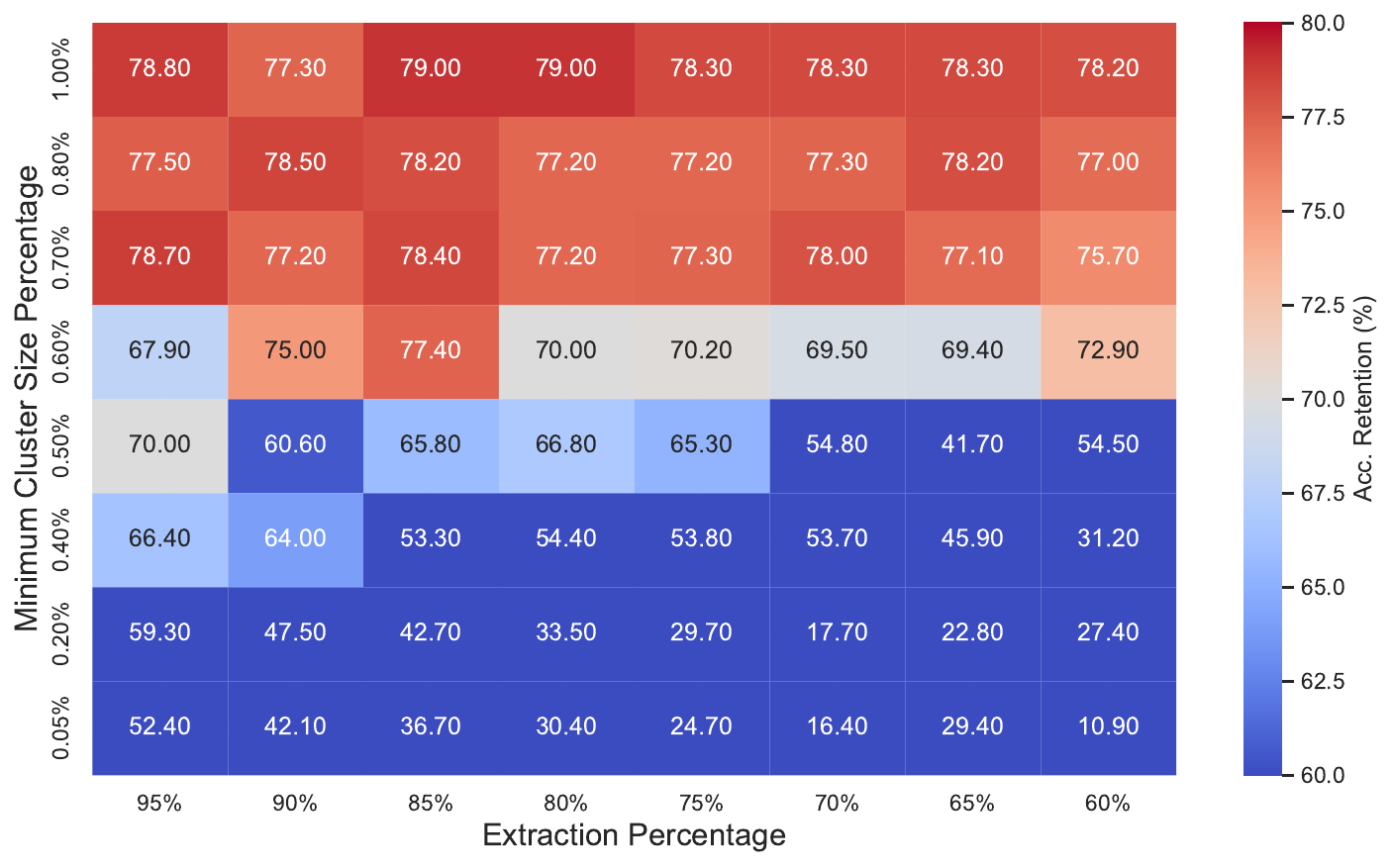}
		\caption{Accuracy Retention After Expert Extraction on \gls{imagenet}: The retention decreases significantly as the minimum cluster size percentage reduces below 0.6\%.}
		\label{appendix:fig:acc-ret}
	\end{subfigure}
	\hfill
	
	\begin{subfigure}[t]{0.48\textwidth}
		\centering
		\includegraphics[width=\textwidth]{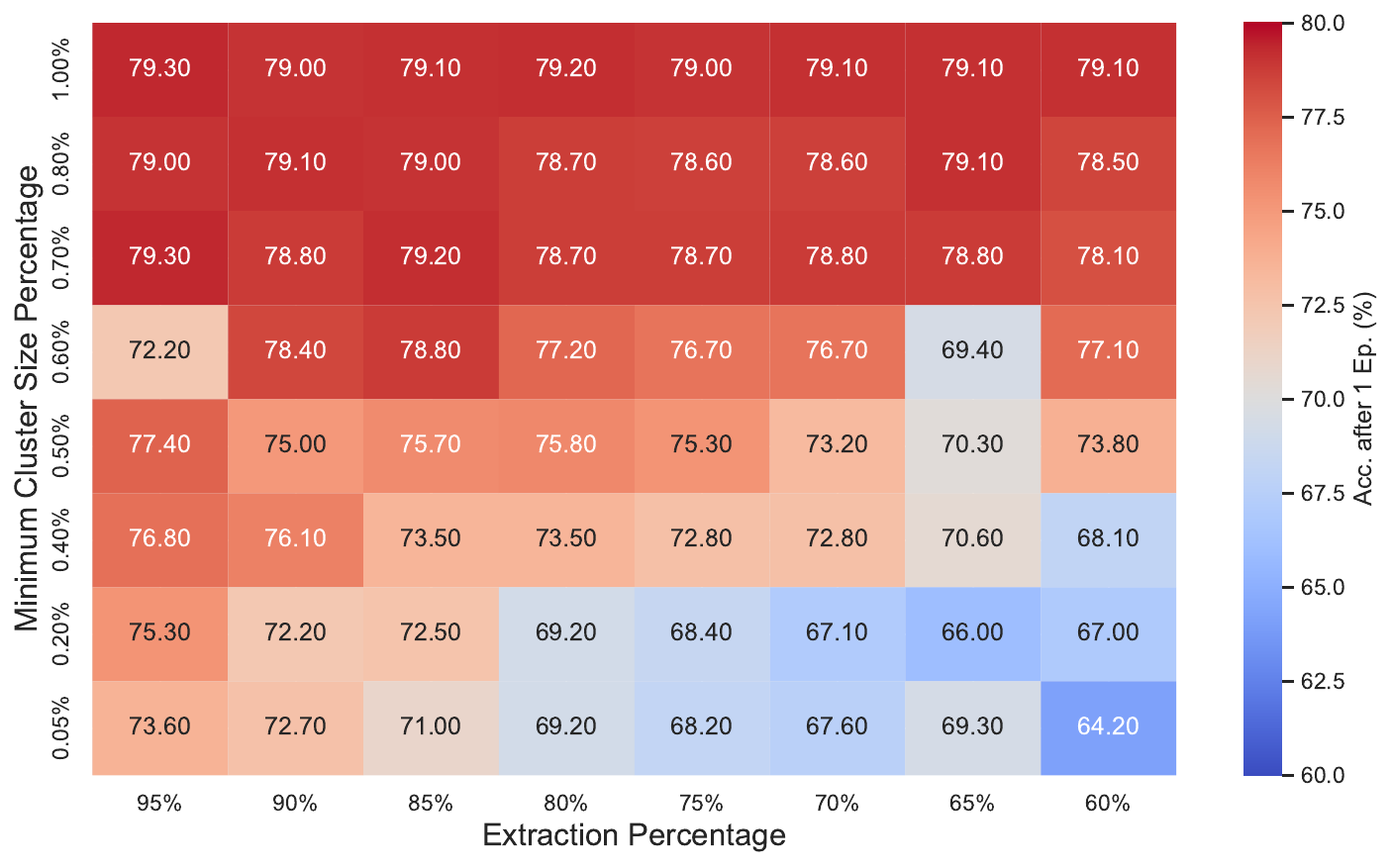}
		\caption{Accuracy After One Epoch of Fine-Tuning on \gls{imagenet}: The accuracy saturates at higher extraction percentages, with minimal gains for increasing the cluster size percentage beyond 0.6\%.}
		\label{appendix:fig:acc-ep1}
	\end{subfigure}
	
	\caption{Heatmaps as a function of extraction percentage and minimum cluster size: (a) immediately following expert extraction and (b) after a single fine-tuning epoch. A clear gradient is observed from bottom-right to top-left in both plots. Fine-tuning mitigates the impact of smaller cluster sizes, with saturation achieved at minimum cluster size percentages above 0.6\%.}
	\label{appendix:fig:comparison-accs}
\end{figure}

\newpage

\begin{figure}[t]
	\centering
	
	\begin{subfigure}[t]{0.48\textwidth}
		\centering
		\includegraphics[width=\textwidth]{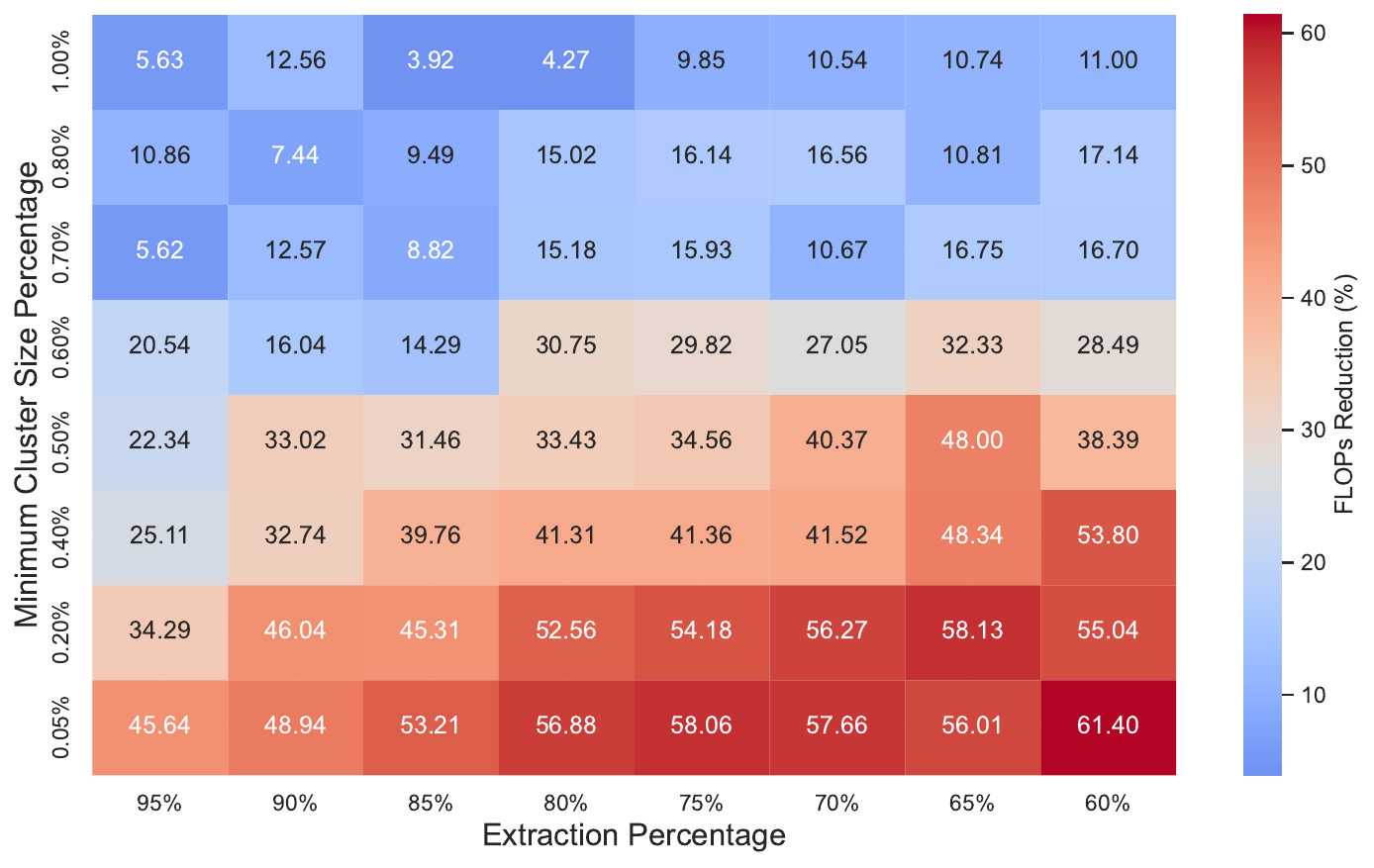}
		\caption{MACs Reduction: The results demonstrate an increase in MACs Reduction with lower minimum cluster sizes and lower extraction percentages.}
		\label{appendix:fig:MACs-red}
	\end{subfigure}
	\hfill
	
	\begin{subfigure}[t]{0.48\textwidth}
		\centering
		\includegraphics[width=\textwidth]{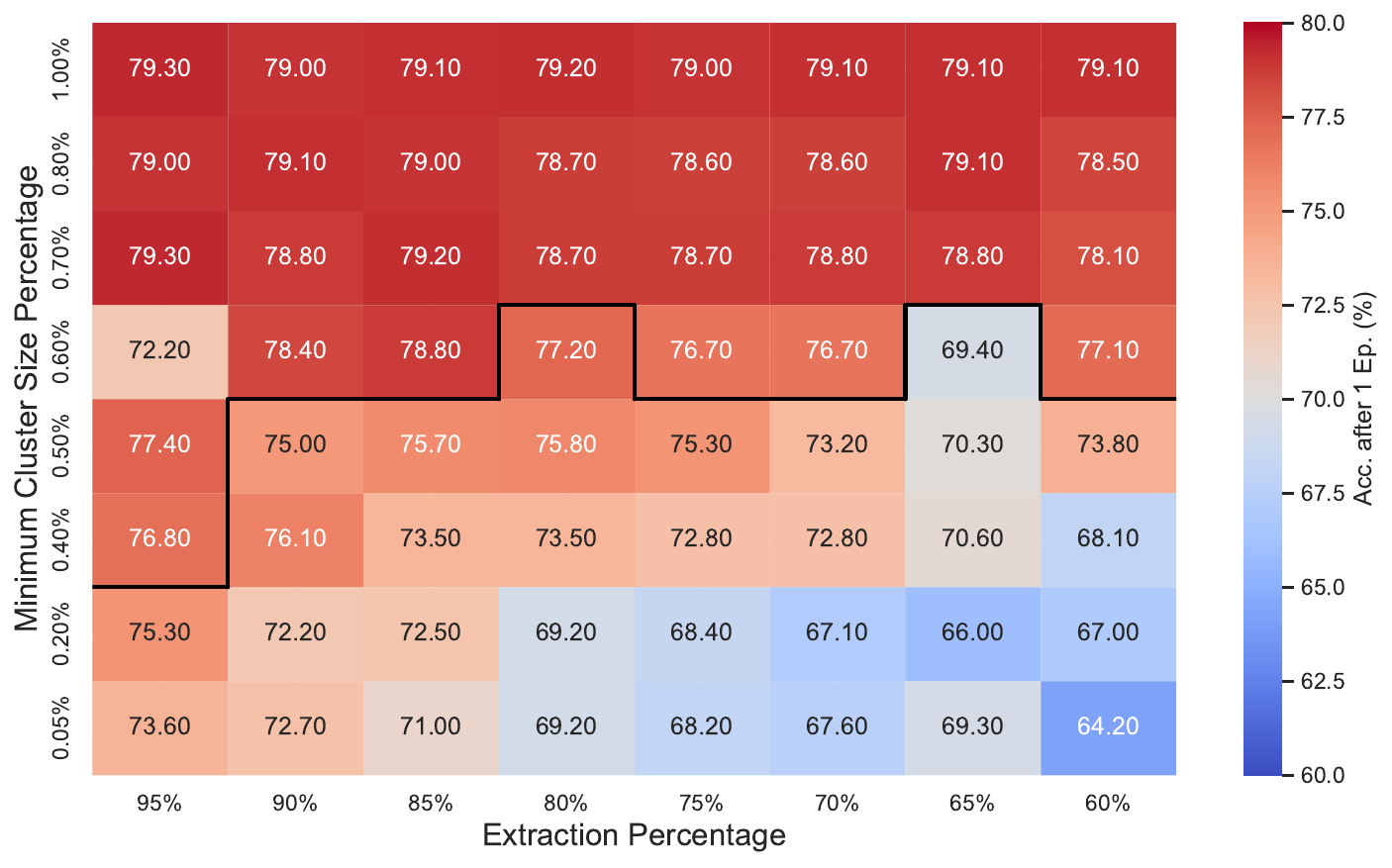}
		\caption{Accuracy After One Epoch of Fine-Tuning on \gls{imagenet}: The threshold highlights the region achieving significant computational efficiency gains.}
		\label{appendix:fig:acc-ep1-line}
	\end{subfigure}
	
	\caption{Heatmaps as a function of extraction percentage and minimum cluster size: (a) MACs Reduction and (b) Top-1 Accuracy after a single fine-tuning epoch. The accuracy decreases as the MACs Reduction increases, highlighting the trade-off between computational efficiency and model performance. This suggests selecting the hyperparameters at the boundary where acceptable accuracy and computational efficiency intersect.}
	\label{appendix:fig:MACs-red-acc-ratio}
\end{figure}

\newpage

\section{Sensitivity to Hyperparameters}
\label{sec:effect-on-hyperparam}

Since we use a subset of the dataset as samples for expert extraction, to select an appropriate minimum cluster size for different sample sizes, we consider the minimum cluster size relative to the total number of samples as a percentage. Increasing this \textit{minimum cluster size percentage} reduces the number of distinct experts because variations in token density are increasingly interpreted as noise within larger clusters by HDBSCAN. Since this reduction in the number of clusters corresponds directly to a reduction in the number of experts, this results in less specialized experts. Therefore, more neurons per expert are required, leading to improved accuracy but lower computational savings. Similarly, higher \textit{extraction percentage} values yield larger experts by preserving more neurons, further enhancing accuracy but at the cost of reduced computational savings.

To select appropriate hyperparameters in this inverse relationship between computational efficiency and accuracy, we perform a hyperparameter search and show our results using heatmaps. 
\\
We use the DeiT-S model evaluated on \gls{imagenet} and apply the same training settings as described in \Cref{sec:experiment}.

\Cref{appendix:fig:comparison-accs} confirms the effect of our hyperparameters, on model accuracy. \Cref{appendix:fig:acc-ret} shows the Accuracy Retention immediately after expert extraction (without fine-tuning), while \Cref{appendix:fig:acc-ep1} shows the accuracy after one epoch of fine-tuning. Both heatmaps show a clear gradient from the bottom-right to the top-left, indicating that larger cluster sizes and higher extraction percentages improve Accuracy Retention.
These results also highlight the saturating effect of fine-tuning. After a single fine-tuning epoch, accuracy increases significantly for smaller cluster sizes, while cluster size percentages above 0.6\% stagnate as they approach the baseline accuracy of $79.7\%$.  
Selecting our hyperparameters near this saturation threshold avoids significant accuracy drops while allowing fine-tuning to improve performance effectively. This suggests that hyperparameters can be selected either immediately after expert extraction or after a single epoch of fine-tuning, making the hyperparameter search simple and computationally efficient.

Furthermore, we aim not only to achieve baseline accuracy but also to achieve a significant reduction in MACs.
For the hyperparameter selection, we therefore need to find the boundary region where computational efficiency and accuracy intersect.
\Cref{appendix:fig:MACs-red-acc-ratio} shows this trade-off between the computational efficiency and model performance. \Cref{appendix:fig:MACs-red} shows the MACs Reduction heatmap and \Cref{appendix:fig:acc-ep1-line} again shows the accuracy after one epoch of fine-tuning, this time with a boundary higlighting a region that achieves a significant reduction in MACs, while maintaining a high accuracy.
For our experiments, we select parameters along this threshold. As seen in \Cref{appendix:fig:acc-ep1-line}, the most top-left point along this boundary corresponds to the best accuracy of $77.2\%$ after a single fine-tuning epoch and a MACs Reduction of $30.75\%$. The selected hyperparameters for our experiments are thus a \textit{minimum cluster size percentage} of 0.6\% and an \textit{extraction percentage} of 80\%.

\section{Sensitivity to Sample Size}
\label{sec:sample-size}

\begin{table}[t] 
	\centering
	\begin{tabularx}{\columnwidth}{
			>{\centering\arraybackslash}X  
			>{\centering\arraybackslash}X  
		}
		\toprule
		\textbf{Number of Input Images} & \textbf{Top-1 Acc. (\%)} \\ 
		\midrule
		320  & 77.93 $\pm$ 2.281 \\
		640  & 78.23 $\pm$ 0.981 \\
		960  & 78.70 $\pm$ 0.529 \\
		1,280 & 78.67 $\pm$ 0.306 \\
		1,600 & 78.93 $\pm$ 0.404 \\
		1,920 & 79.00 $\pm$ 0.100 \\
		2,240 & 79.37 $\pm$ 0.153 \\
		2,560 & 79.00 $\pm$ 0.173 \\
		\bottomrule
	\end{tabularx}
	\caption{Mean \textbf{Top-1 Accuracy (\%)} of DeiT-S on \gls{imagenet} as a function of the \textbf{Number of Input Images} used for the expert extraction. Each accuracy is presented alongside its standard deviation to demonstrate the variability. For a higher number of input images the standard deviation drops, indicating that the extraction procedure stabilizes.}
	\label{tab:mean-top1-accuracy-images}
\end{table}

To evaluate the effect of the sample size on our method, we use the DeiT-S model, applying the same training settings as described in \Cref{sec:experiment} of the paper.

\Cref{tab:mean-top1-accuracy-images} presents the mean Top-1 Accuracy on \gls{imagenet} over three random seeds, along with the standard deviation, for different numbers of input images used to identify clusters during expert extraction. The results indicate a clear trend: as the number of input samples increases, the standard deviation decreases. However, this improvement comes with a computational trade-off due to the non-linear runtime complexity of the HDBSCAN clustering algorithm with respect to the number of samples.

For our experiments, we select 640 input images, corresponding to 126,080 sample tokens clustered per layer. This configuration achieves a standard deviation below 1\%, balancing runtime and accuracy consistency.
Notably, this stabilization of the standard deviation and accuracy occurs at a number of input images (640) that cannot even represent each class in the \gls{imagenet} dataset (1,000 classes). We attribute this to the fact that the clustering is based on tokens derived from image patches, which are redundant and shared across multiple classes, as seen in \Cref{further-analysis}. This redundancy in token distributions allows for robust clustering even with fewer images than the total number of classes.

\section{Comparison to MoEfication}

\label{sec:comparison-to-moefication}

\begin{table}[t]
	\centering
	\small
	\renewcommand{\arraystretch}{1.2} 
	\begin{tabular}{
			>{\raggedright\arraybackslash}p{2.5cm}  
			>{\centering\arraybackslash}p{2.25cm} 
			>{\centering\arraybackslash}p{2.25cm}
		}
		\toprule
		\textbf{Aspect} & \textbf{MoEfication} & \textbf{Ours} \\
		\midrule
		Expert Count    & Manually Chosen     & Data-Driven \\
		\midrule
		Expert Structure& Disjoint Partitions & Overlapping Subnetworks  \\
		\midrule
		Routing Target  & Mean Weight Columns & Mean Input Tokens \\
		\midrule
		Domain Focus    & Large Language Models & Vision Transformers \\
		\bottomrule
	\end{tabular}
	\caption{Comparison of our method to MoEfication~\cite{mtfflamoe}.}
	\label{tab:comparison-to-moefication}
\end{table}

Zhang et al.~\cite{mtfflamoe} rely on weight co-activation graphs and a manually set number of experts. In contrast, we cluster activations and let the data determine the number of experts automatically. Their disjoint partitioning contrasts with our variance-based extraction that allows overlapping experts, leading to fewer constraints on the experts. Moreover, while \cite{mtfflamoe} routes tokens to the most similar mean weight column, we compute similarities in the input space directly (see \Cref{tab:comparison-to-moefication}). Our results on \gls{imagenet} validate this design, showing performance gains in vision tasks.

\begin{table*}[t]  
	\centering
	\begin{tabularx}{\textwidth}{
			>{\arraybackslash}p{4.0cm}
			>{\centering\arraybackslash}X
			>{\centering\arraybackslash}X
			>{\centering\arraybackslash}p{3.0cm}
			>{\centering\arraybackslash}X
		}
		\toprule
		\textbf{Model} & \textbf{MACs (G)} & \textbf{Parameters (M)} & \textbf{Acc. Retention (\%)} & \textbf{Top-1 Acc. (\%)} \\
		\midrule
		\rowcolor{gray!20} Swin-T & 4.50 & 28.29 & -- & 81.19 \\
		Swin-T-MoEE (Ours) & 3.59 (-20.2\%) & 17.17 (-39.3\%) & 58.14 & 79.02 \\
		\rowcolor{gray!20} Swin-S & 8.76 & 49.61 & -- & 83.20 \\
		Swin-S-MoEE (Ours) & 6.19 (-29.4\%) & 30.65 (-38.2\%) & 51.16 & 81.21 \\
		\rowcolor{gray!20} Swin-B & 15.46 & 87.77 & -- & 83.47 \\
		Swin-B-MoEE (Ours) & 10.42 (-32.6\%) & 53.05 (-39.6\%) & 50.17 & 83.12 \\
		\midrule
		\rowcolor{gray!20} ConvNeXt-T & 4.47 & 28.59 & -- & 82.12 \\
		ConvNeXt-T-MoEE (Ours) & 3.53 (-21.0\%) & 18.57 (-35.0\%) & 35.28 & 81.82 \\
		\rowcolor{gray!20} ConvNeXt-S & 8.71 & 50.22 & -- & 83.11 \\
		ConvNeXt-S-MoEE (Ours) & 6.83 (-21.6\%) & 34.00 (-32.3\%) & 34.54 & 82.68 \\
		\rowcolor{gray!20} ConvNeXt-B & 15.38 & 88.59 & -- & 83.80 \\
		ConvNeXt-B-MoEE (Ours) & 10.85 (-29.5\%) & 49.16 (-44.5\%) & 34.81 & 83.23 \\
		\bottomrule
	\end{tabularx}
	
	\caption{Performance and parameter comparison for additional architectures, evaluated using: \textbf{Accuracy Retention}: retained accuracy after expert extraction, before fine-tuning; \textbf{Top-1 Accuracy}: final accuracy after fine-tuning; \textbf{MACs}: computational operations, measured in billions of operations; and \textbf{Parameters}: the total model size in millions of parameters. Our method (MoEE) generalizes to other architectures by achieving competitive accuracy with significant reductions in MACs and parameters, especially in the Swin-S, Swin-B and ConvNeXt-B models.}
	
	\label{tab:architectures}
\end{table*}

\begin{table*}[t]  
	\centering
	\begin{tabularx}{\textwidth}{
			>{\arraybackslash}p{4.0cm}
			>{\centering\arraybackslash}X
			>{\centering\arraybackslash}X
			>{\centering\arraybackslash}p{3.0cm}
			>{\centering\arraybackslash}X
		}
		\toprule
		\textbf{Model} & \textbf{MACs (G)} & \textbf{Parameters (M)} & \textbf{Acc. Retention (\%)} & \textbf{Top-1 Acc. (\%)} \\ 
		\midrule
		
		\rowcolor{gray!20} DeiT-T & 1.30 & 5.72 & -- & 80.50 \\
		DeiT-T-MoEE (Ours) & 1.03 (-20.8\%) & 4.56 (-20.3\%) & 10.24 & 79.32 \\
		\rowcolor{gray!20} DeiT-S & 4.61 & 22.05 & -- & 85.33 \\
		DeiT-S-MoEE (Ours) & 3.55 (-23.0\%) & 16.70 (-24.3\%) & 55.85 & 84.94 \\
		\rowcolor{gray!20} DeiT-B & 17.58 & 86.57 & -- & 88.20 \\
		DeiT-B-MoEE (Ours) & 11.73 (-33.3\%) & 56.83 (-34.4\%) & 15.76 & 86.71 \\
		
		\bottomrule
	\end{tabularx}
	
	\caption{Performance and parameter comparison of the DeiT-MoEE models, evaluated on CIFAR-100 using:  \textbf{Accuracy Retention}: retained accuracy after expert extraction, before fine-tuning; \textbf{Top-1 Accuracy}: final accuracy after fine-tuning; \textbf{MACs}: computational operations, measured in billions of operations; and \textbf{Parameters}: the total model size in millions of parameters. Our method (MoEE) generalizes to smaller datasets and thus data efficient settings.}
	
	\label{tab:cifar100}
\end{table*}

\newpage

\section{Generalizability to other Datasets and Architectures}
\label{sec:generalizability}

The results in \Cref{tab:architectures} and \Cref{tab:cifar100} confirm that our method generalizes well across both hierarchical and convolution-inspired transformer architectures (Swin-Transformer \citep{sthvtusw} and ConvNeXt Models \citep{acft2}), as well as to smaller datasets like CIFAR-100 \citep{lmloffti}. Despite structural differences, all variants benefit from fewer MACs, reduced parameter counts and competitive final accuracies.

In particular, the larger base variants benefit most from expert extraction. For Swin-B-MoEE and ConvNeXt-B-MoEE, we reduce the MACs by $33\%$ and $30\%$, and the parameter count by $40\%$ and $45\%$ respectively, while maintaining over $98\%$ of the original accuracy after fine-tuning. These results confirm that our method becomes increasingly effective with model size, offering substantial savings in both compute and memory without compromising final performance.
\\
\\
Notably, we observe that the parameter reduction is more pronounced in models like Swin and ConvNeXt compared to DeiT. Which we primarily attribute to their hierarchical structure. Since experts predominantly form in the deeper layers, where token throughput is lower but embedding dimensions are larger, this results in higher parameter savings but lower relative MAC reductions.

Furthermore, ConvNeXt models show notably lower accuracy retention before fine-tuning. We attribute this to their use of convolutions, which are inherently more parameter-efficient than fully connected layers. Thus, removing even a small set of neurons can have a stronger relative effect on expressiveness. Nevertheless, the method still recovers strong final accuracy after fine-tuning, showing that even architectures with different inductive biases remain compatible with our approach.

\begin{table*}[t]  
	\centering
	\begin{tabularx}{\textwidth}{
			>{\arraybackslash}X
			>{\centering\arraybackslash}X
			>{\centering\arraybackslash}X
			>{\centering\arraybackslash}X
			>{\centering\arraybackslash}X
		}
		\toprule
		\textbf{Model} & \textbf{MACs (G)} & \textbf{Parameters (M)} & \textbf{Acc. Retention (\%)} & \textbf{Top-1 Acc. (\%)} \\ 
		\midrule
		\rowcolor{gray!20} DeiT-S & 4.61 & 22.05 & -- & 79.70 \\
		HDBSCAN (Ours) & \textbf{3.19 (-30.6\%)} & 16.54 (-25.0\%) & \textbf{67.60} & \textbf{78.11} \\
		DBSCAN & 3.52 (-23.5\%) & 16.53 (-25.1\%) & 66.47 & 77.70 \\
		OPTICS & 3.60 (-21.7\%) & 16.92 (-23.3\%) & 67.06 & 77.79 \\
		K-Means & 3.54 (-23.1\%) & 16.60 (-24.7\%) & 47.34 & 76.68 \\
		BIRCH & 3.51 (-23.7\%) & \textbf{16.45 (-25.4\%)} & 48.33 & 77.01 \\
		\bottomrule
	\end{tabularx}
	
	\caption{Performance and parameter comparison for different clustering algorithms, evaluated using: \textbf{Accuracy Retention}: retained accuracy after expert extraction, before fine-tuning; \textbf{Top-1 Accuracy}: final accuracy after fine-tuning; \textbf{MACs}: computational operations, measured in billions of operations; and \textbf{Parameters}: the total model size in millions of parameters. Our method (using HDBSCAN) provides the best performance among the density-based clustering algorithms and significantly outperforms the partition-based algorithms.}
	
	\label{tab:clustering}
\end{table*}

\begin{table*}[t]
	\centering
	\begin{tabularx}{\textwidth}{
			>{\arraybackslash}X
			>{\centering\arraybackslash}X
			>{\centering\arraybackslash}X
			>{\centering\arraybackslash}X
			>{\centering\arraybackslash}X
			>{\centering\arraybackslash}X
			>{\centering\arraybackslash}X
		}
		\toprule
		\textbf{Method} & \multicolumn{3}{c}{\textbf{Magnitude-Based}} & \multicolumn{3}{c}{\textbf{Variance-Based}} \\
		\cmidrule(lr){2-4} \cmidrule(lr){5-7}
		& \textbf{MACs Reduction (\%)} & \textbf{Acc. Retention (\%)} & \textbf{Top-1 Acc. (\%)} 
		& \textbf{MACs Reduction (\%)} & \textbf{Acc. Retention (\%)} & \textbf{Top-1 Acc. (\%)} \\
		\midrule
		\textbf{Cosine} & 28.17 & 64.45 & 77.15 & 30.63 & 67.60 & 78.20 \\
		\textbf{Euclidean} & 27.22 & 63.90 & 77.35 & 30.48 & 66.15 & 78.10 \\
		\midrule
		\textbf{Hashing} & 22.34 & 40.92 & 76.75 & 24.98 & 46.53 & 77.11 \\
		\bottomrule
	\end{tabularx}
	
	\caption{Comparison of routing methods (\textbf{Cosine Similarity}, \textbf{Euclidean Distance} and \textbf{Hashing}) using two extraction strategies (\textbf{Magnitude-Based} and \textbf{Variance-Based}) on DeiT-S and evaluated on \gls{imagenet}. The table evaluates three metrics: \textbf{Accuracy Retention (\%)}: retained accuracy after expert extraction before fine-tuning; \textbf{Top-1 Accuracy (\%)}: final accuracy after fine-tuning; and \textbf{MACs Reduction (\%)}: relative computational savings compared to the baseline model. Variance-Based extraction consistently outperforms Magnitude-Based extraction for both routing methods, achieving higher Accuracy Retention, Top-1 Accuracy, and MACs Reduction. Between the routing methods there is no significant difference, as both Cosine Similarity and Euclidean Distance show comparable results across all metrics. These results highlight the robustness of the extraction strategies regardless of the used routing method, with Variance-Based approaches generally providing better performance.}
	
	\label{tab:input_routing_comparison}
\end{table*}

\section{Effect of Clustering Algorithm}
\label{sec:effect-of-clustering}

To evaluate impact of different clustering algorithms on expert extraction, we conduct experiments on the DeiT-S model using the same training settings as described in \Cref{sec:experiment} and evaluate the resulting model on \gls{imagenet}. 

We compare our default method, HDBSCAN, against other density-based methods (DBSCAN, OPTICS) as well as partition-based alternatives (K-Means, BIRCH). For the partition-based algorithms, which require the number of clusters as a hyperparameter, we guide their selection using the number of experts extracted by HDBSCAN per layer (see \Cref{tab:nr-experts}). As shown in \Cref{tab:clustering}, density-based methods consistently outperform partition-based ones in the Top-1 Accuracies, particularly in accuracy retention. Among the density-based methods, HDBSCAN achieves the best trade-off across all metrics, leading to the highest accuracy retention and final accuracy, while also yielding the greatest reduction in MACs. These results justify our choice of HDBSCAN for all main experiments in this work.

\begin{figure*}[t]
	\centering
	\begin{subfigure}[t]{0.48\textwidth}
		\centering
		\includegraphics[width=\textwidth]{l11_truck_routings.pdf}
		\caption{Token routing distributions at different layers for visually similar truck-like classes (fire engine, garbage truck, pickup, and tow truck) compared to the distribution across all \gls{imagenet} classes (all). These classes show similar routing patterns among themselves and align with the distribution of all classes.}
		\label{appendix:fig:truck-routings-11}
	\end{subfigure}
	\hfill
	\begin{subfigure}[t]{0.48\textwidth}
		\centering
		\includegraphics[width=\textwidth]{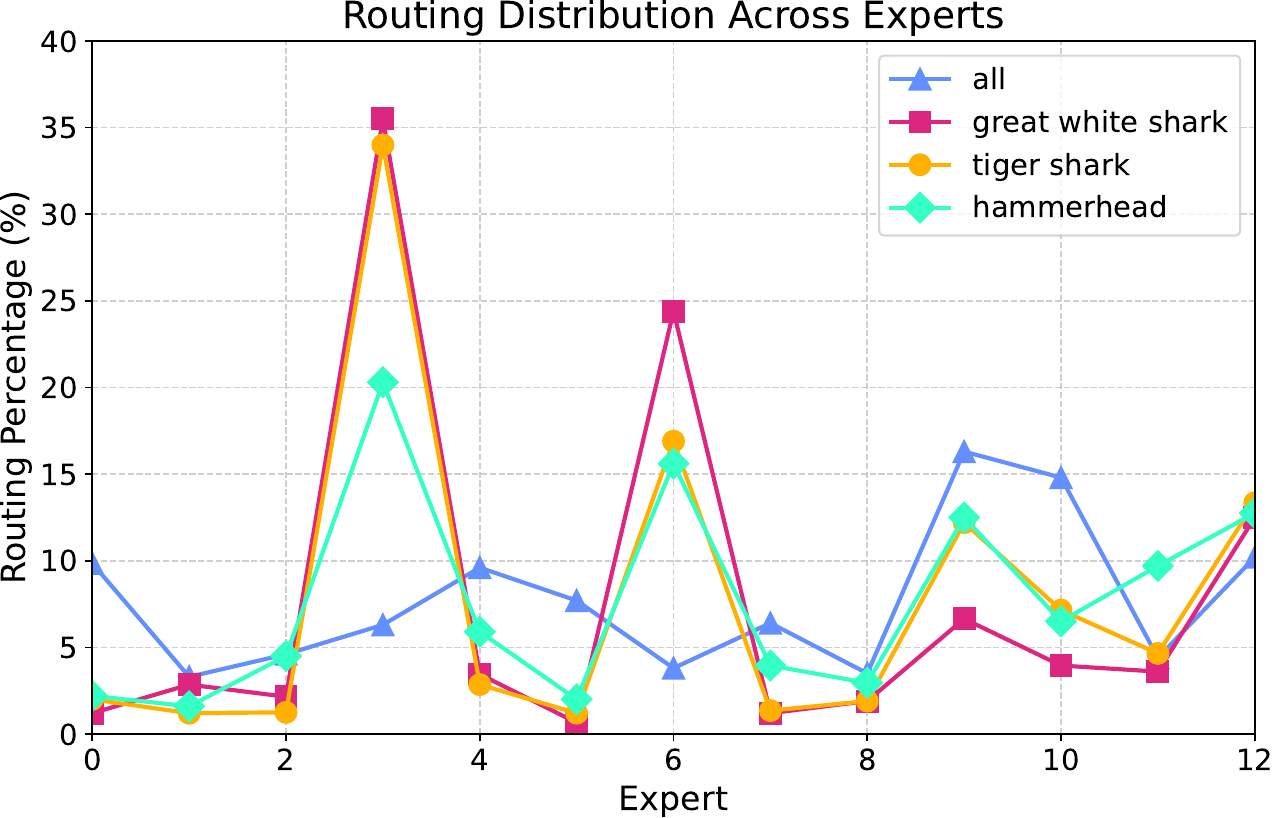}
		\caption{Token routing distributions at different layers for visually similar shark-like classes (great white shark, tiger shark, and hammerhead) compared to the distribution across all \gls{imagenet} classes (all). These classes show similar routing patterns among themselves but differ significantly from the distribution of all classes.}
		\label{appendix:fig:shark-routings-11}
	\end{subfigure}
	\caption{Comparison of token routing distributions for visually similar classes in layer 11. (a) presents routing patterns for truck-like classes, which closely resemble the overall routing distribution of all classes. (b) shows routing patterns for shark-like classes, which form similar patterns that diverge from the overall distribution. These results suggest that truck-like classes result in more generic tokens shared across multiple classes.}
	\label{appendix:fig:comparison-routings}
\end{figure*}

\section{Effect of Extraction and Routing Method}
\label{sec:effect-of-extraction-and-routing}

In order to evaluate the effects of the extraction and routing methods, we perform another ablation study. For the extraction strategy, we consider two methods for selecting the hidden neurons of each expert. In the Magnitude-Based approach, neurons are selected based on their mean activation magnitude, prioritizing neurons with higher average activations. The Variance-Based approach instead prioritizes neurons with higher within-cluster variance, capturing diversity in activation patterns.

Additionally, we compare three routing approaches: Cosine Similarity, where new input tokens are routed to the expert with the highest cosine similarity; Euclidean Distance, which selects the expert with the smallest Euclidean distance to the cluster mean; and Hashing, an orthogonal method that uses a hash function for routing. Unlike Cosine and Euclidean routing, which assume a Gaussian cluster shape and rely on descriptive statistics of the inputs, hashing requires that the same function be used during extraction and inference.
\\

\begin{figure}[t]
	\centering
	\includegraphics[width=\columnwidth]{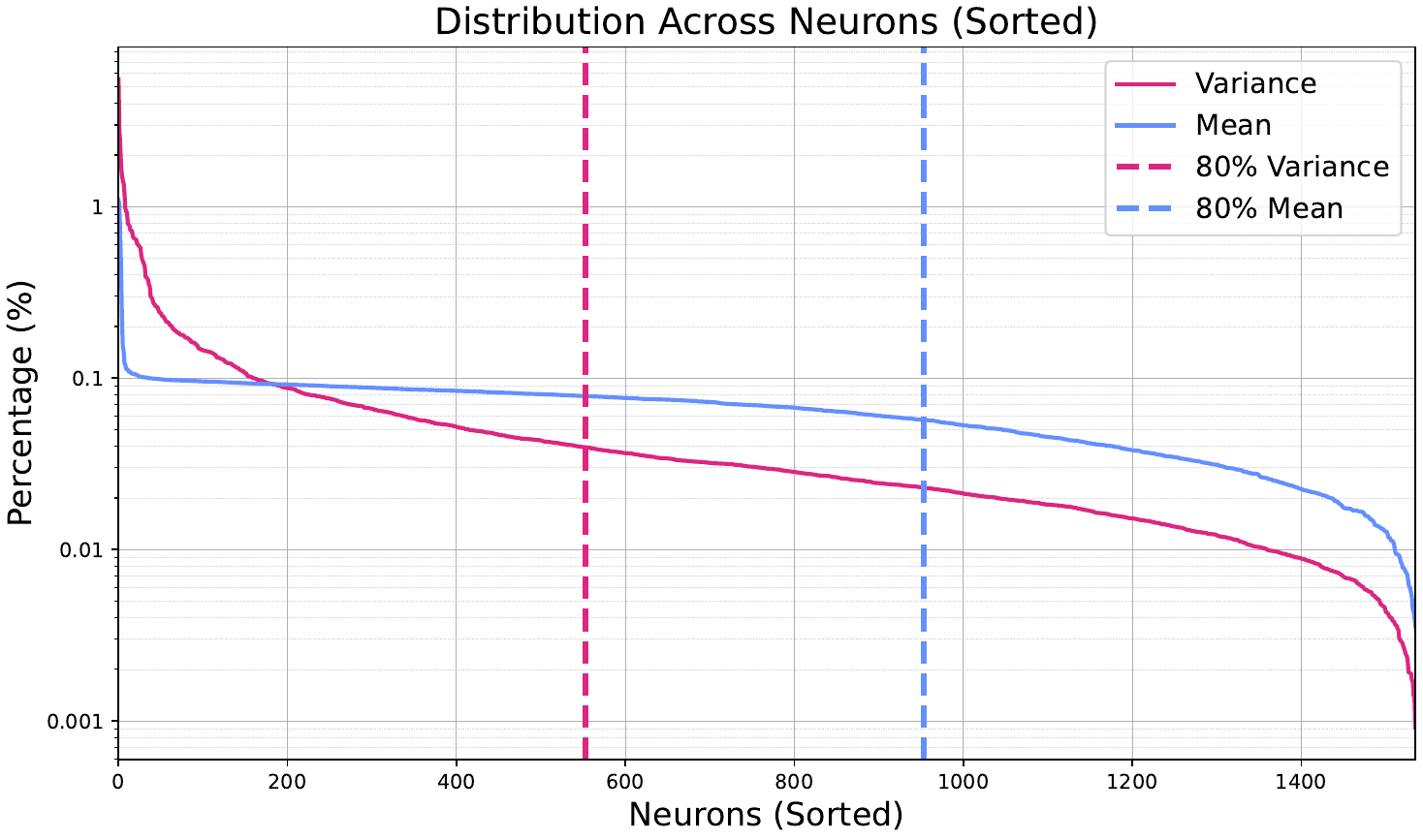}
	\caption{Distribution of variance and mean across neurons in layer 11. The y-axis uses a logarithmic scale to emphasize the differences between neurons. Vertical dashed lines indicate the points where the cumulative sum of the variance and mean reaches 80\%. These results demonstrate that the mean is more uniformly distributed, requiring a larger number of neurons to achieve the same cumulative percentage compared to the variance.}
	\label{fig:neuron-distribution-log}
\end{figure}

Notably, Magnitude-Based prioritization results in a smaller reduction in MACs for a given extraction percentage compared to Variance-Based prioritization. As shown in \Cref{fig:neuron-distribution-log}, this is because of differences in the distributions of variance and mean across neurons. The mean activations have a more uniform distribution, necessitating a larger number of neurons to account for a specific cumulative percentage. 
\\
\\
In contrast, the variance is concentrated among fewer neurons, allowing Variance-Based extraction to cover a larger cumulative percentage with fewer neurons. This concentration results in greater MACs reductions for the same extraction percentage.
To enable a fair comparison between the two methods at similar MACs reductions, the extraction percentage is adjusted from 80\% for Variance-Based extraction to 60\% for Magnitude-Based extraction.

\Cref{tab:input_routing_comparison}, shows that Variance-Based extraction consistently outperforms Magnitude-Based extraction across all metrics: Accuracy Retention, Top-1 Accuracy, and MACs Reduction. 
On the other hand, the choice of routing method had no significant impact on the overall results, with Cosine Similarity and Euclidean Distance having comparable accuracy and computational savings in both extraction scenarios. As the Cosine Similarity can be computed without normalization, we chose the Cosine Similarity as the default routing method in all experiments, due to the better computational efficiency.

\begin{table*}[t]  
	\centering
	\begin{tabularx}{\textwidth}{
			>{\arraybackslash}p{2.5cm}  
			>{\centering\arraybackslash}p{1.5cm}
			>{\centering\arraybackslash}X
			>{\centering\arraybackslash}X
			>{\centering\arraybackslash}X
			>{\centering\arraybackslash}X
			>{\centering\arraybackslash}X
			>{\centering\arraybackslash}X
			>{\centering\arraybackslash}X
			>{\centering\arraybackslash}X
			>{\centering\arraybackslash}X
			>{\centering\arraybackslash}X
			>{\centering\arraybackslash}X
		}
		\toprule
		\textbf{Model}         & \textbf{Layer 0} & \textbf{ 1} & \textbf{2}  & \textbf{ 3}  & \textbf{ 4}  & \textbf{ 5} & \textbf{ 6} & \textbf{ 7} & \textbf{ 8} & \textbf{ 9} & \textbf{ 10} & \textbf{ 11} \\ 
		\midrule
		DeiT-T-MoEE  & -- & -- & -- & -- & -- & -- & --  & 4.67 & 7.67 & 8.33 & 8.00 & 10.33 \\
		DeiT-S-MoEE  & -- & -- & -- & -- & -- & -- & 2.67 & 6.00 & 7.67 & 8.67 & 8.67 & 10.67 \\
		DeiT-B-MoEE  & -- & -- & -- & -- & -- & -- & 7.67 & 8.00 & 9.00 & 8.33 & 9.00 & 11.00 \\
		
		\bottomrule
	\end{tabularx}
	\caption{Mean number of experts extracted across layers for different DeiT-MoEE models. Specialization into distinct activation clusters emerges progressively in the deeper layers.}

	\label{tab:nr-experts}
\end{table*}

\begin{table*}[t]
	\centering
	\begin{tabularx}{\textwidth}{
			>{\arraybackslash}p{3cm}  
			>{\centering\arraybackslash}X  
			>{\centering\arraybackslash}X  
			>{\centering\arraybackslash}X  
			>{\centering\arraybackslash}X  
		}
		\toprule
		\textbf{Model}         & \textbf{Stage 0} & \textbf{Stage 1} & \textbf{Stage 2} & \textbf{Stage 3} \\
		\midrule
		Swin-T-MoEE   & --   & --   & 2.00 & 5.50 \\
		Swin-S-MoEE   & --   & --   & 2.10 & 9.00 \\
		Swin-B-MoEE   & --   & --   & 2.90 & 10.50 \\
		\midrule
		ConvNeXt-T-MoEE & -- & -- & 2.00 & 4.50 \\
		ConvNeXt-S-MoEE & -- & -- & 2.50 & 4.50 \\
		ConvNeXt-B-MoEE & -- & -- & 2.15 & 6.00 \\
		\bottomrule
	\end{tabularx}
	\caption{Mean number of experts activated across model stages for Swin and ConvNeXt MoEE variants. Expert specialization emerges predominantly in the later stages, especially in deeper variants.}
	\label{tab:nr-experts-other}
\end{table*}

\section{Insights into Routing Distributions}
\label{further-analysis}

We further analyse the routing distribution in \Cref{appendix:fig:comparison-routings}, which shows the token routing distributions for selected visually similar classes in layer 11, compared to the distribution across all 1,000 \gls{imagenet} classes. Specifically, \Cref{appendix:fig:truck-routings-11} presents the routing distributions for truck-like classes (fire engine, garbage truck, pickup, and tow truck), while \Cref{appendix:fig:shark-routings-11} shows the distributions for shark-like classes (great white shark, tiger shark, and hammerhead).
\\
The routing patterns of visually similar classes show a high degree of overlap, suggesting similar classes are processed through similar expert selections. However, the routing distributions between truck-like and shark-like classes are noticeably different from each other, indicating that visual similarity influences token routing, even after the positional encodings have modified the token representations.

Furthermore, the distribution for the truck-like classes closely aligns with the distribution of all 1,000 \gls{imagenet} classes, unlike the shark-like classes. These show cohesive routing patterns among themselves but differ significantly from the overall distribution. Since the routing is based on a Cosine Similarity to fixed mean tokens, this similarity in routing distributions means that tokens in truck-like classes are similar to tokens across all classes. This again confirms that the token redundancy allows the model to generalize routing patterns effectively (see \Cref{sec:sample-size}).

\section{Insights into Expert Formations}

\Cref{tab:nr-experts} and \ref{tab:nr-experts-other} show the mean number of experts extracted in each layer for different models. Notably, the earlier layers do not exhibit any formed clusters, reflecting the more general feature representations in shallower layers. Deeper layers, on the other hand, display more experts reflecting the progressively stronger specialization, which aligns with our analysis that more discriminative or class-relevant features emerge later in the network.

An important aspect is that each experts can vary in size and may partially overlap in their selected neurons. This partial overlap means that a strict one-to-one partition of the MLP into disjoint experts does not necessarily occur. Nonetheless, the union of all extracted experts in a given layer remains smaller than the original MLP of that layer, as evidenced by the reduced parameter counts in our results.

\begin{figure}[t] 
	
	\begin{subfigure}[t]{\columnwidth}
		\centering
		\includegraphics[width=\columnwidth]{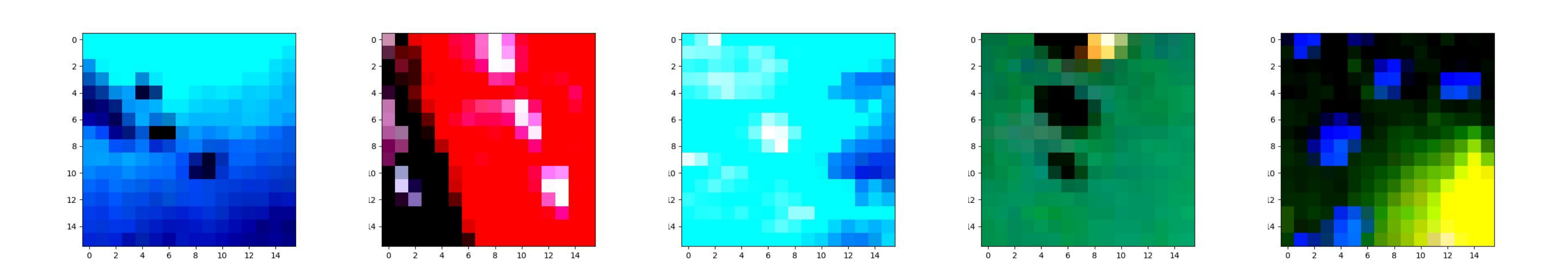}
		\caption{Selected sample patches routed to expert 3 in layer 11.}
		\label{fig:expert3}
	\end{subfigure}
	
	\begin{subfigure}[t]{\columnwidth}
		\centering
		\includegraphics[width=\columnwidth]{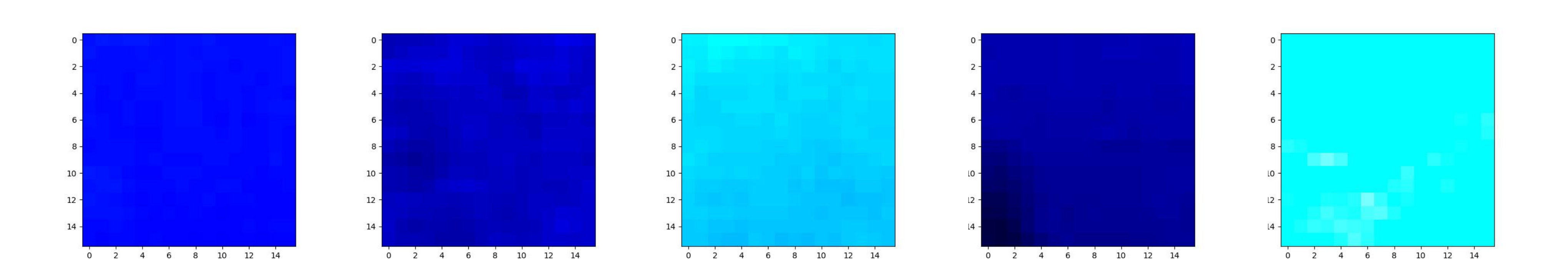}
		\caption{Selected sample patches routed to expert 6 in layer 11.}
		\label{fig:expert6}
	\end{subfigure}
	
	\begin{subfigure}[t]{\columnwidth}
		\centering
		\includegraphics[width=\columnwidth]{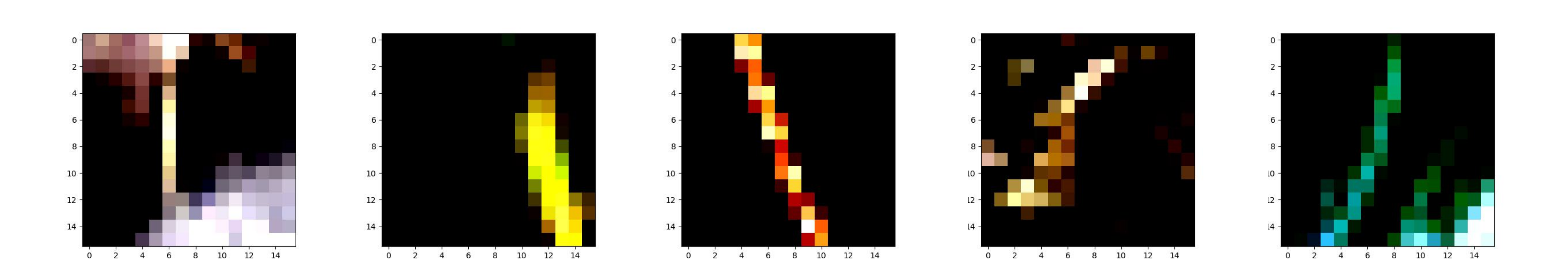}
		\caption{Selected sample patches routed to expert 10 in layer 11.}
		\label{fig:expert10}
	\end{subfigure}

	\caption{Figure of selected sample patches from DeiT-B}	
	\label{fig:patches}
\end{figure}

To provide a qualitative analysis, we additionally visualize the image patches corresponding to tokens routed to specific experts. We first save the token routings for the selected experts at layer 11 for random batches of validation data. We then identify the corresponding image patches from the original input for each token. Note that these patches have undergone multiple processing steps in MHSA and MLP layers, so the tokens may no longer resemble the original patches.
For a better insight, the resulting patches are then clustered, and samples within one cluster are selected for visualization, providing a visual understanding of the type of patches routed to each expert (see \Cref{fig:patches}).